\documentclass[journal]{IEEEtran}
\usepackage{cite}
\usepackage{amsmath,amssymb,amsfonts}
\usepackage{bbm}
\DeclareMathOperator*{\argmin}{arg\,min}
\usepackage{gensymb}
\usepackage{algorithm}
\usepackage{algpseudocode}
\algrenewcommand\algorithmicrequire{\textbf{Input:}}
\algrenewcommand\algorithmicensure{\textbf{Output:}}
\usepackage{graphicx}
\usepackage{booktabs}
\usepackage{subcaption}
\usepackage{textcomp}
\usepackage{url}
\usepackage{booktabs}
\usepackage{relsize}
\usepackage{xcolor}
\usepackage{multirow}
\usepackage[hidelinks]{hyperref}
\usepackage{microtype}[final]
\usepackage{pifont}% http://ctan.org/pkg/pifont

\newcommand{\rotninty}[1]{\rotatebox[origin=c]{90}{#1}}

\newcommand{\rev}[1]{\textcolor{black}{#1}}

\newcommand{\behzad}[1]{\textcolor{black}{#1}}

\newcommand{\ours}{\text{Distill-SODA}}

\newcommand{\negmargin}[1]{ {\color{red}{(-#1)}}}
\definecolor{darkgreen}{HTML}{3C8031}
\newcommand{\posmargin}[1]{ {\textcolor{darkgreen}{(+#1)}}}

\begin{document}
\title{Distill-SODA: \underline{Distill}ing Self-Supervised Vision Transformer for \underline{S}ource-Free \underline{O}pen-Set \underline{D}omain \underline{A}daptation in Computational Pathology}
% \title{Source-Free Open-Set Domain Adaptation for Computational Pathology via Distilling Self-Supervised Vision Transformer}
%

\author{\IEEEauthorblockN{Guillaume Vray$^{1}$ \quad
Devavrat Tomar$^{1}$ \quad
Jean-Philippe Thiran$^{1,2}$ \quad
Behzad Bozorgtabar$^{1,2}$ \\
\IEEEauthorblockA{$^{1}$EPFL \hspace{2em} $^{2}$CHUV} \\
\IEEEauthorblockA{Email: \{firstname\}.\{lastname\}@epfl.ch}
% \IEEEauthorblockA{\IEEEauthorrefmark{2}Twentieth Century Fox, Springfield, USA\\
% Email: homer@thesimpsons.com}
% \IEEEauthorblockA{\IEEEauthorrefmark{3}Starfleet Academy, San Francisco, California 96678-2391\\
% Telephone: (800) 555--1212, Fax: (888) 555--1212}
% \IEEEauthorblockA{\IEEEauthorrefmark{4}Tyrell Inc., 123 Replicant Street, Los Angeles, California 90210--4321}
}}
\maketitle
\thispagestyle{plain}
\pagestyle{plain}
\begin{abstract}
Developing computational pathology models is essential for reducing manual tissue typing from whole slide images, transferring knowledge from the source domain to an unlabeled, shifted target domain, and identifying unseen categories. We propose a practical setting by addressing the above-mentioned challenges in one fell swoop, i.e., source-free open-set domain adaptation. Our methodology focuses on adapting a pre-trained source model to an unlabeled target dataset and encompasses both closed-set and open-set classes. Beyond addressing the semantic shift of unknown classes, our framework also deals with a covariate shift, which manifests as variations in color appearance between source and target tissue samples. Our method hinges on distilling knowledge from a self-supervised vision transformer (ViT), drawing guidance from either robustly pre-trained transformer models or histopathology datasets, including those from the target domain. In pursuit of this, we introduce a novel style-based adversarial data augmentation, serving as hard positives for self-training a ViT, resulting in highly contextualized embeddings. Following this, we cluster semantically akin target images, with the source model offering weak pseudo-labels, albeit with uncertain confidence. To enhance this process, we present the closed-set affinity score (CSAS), aiming to correct the confidence levels of these pseudo-labels and to calculate weighted class prototypes within the contextualized embedding space. Our approach establishes itself as state-of-the-art across three public histopathological datasets for colorectal cancer assessment. Notably, our self-training method seamlessly integrates with open-set detection methods, resulting in enhanced performance in both closed-set and open-set recognition tasks.
\end{abstract}


%Thus, these open-set samples can be detected and treated differently by manual expert review preventing unconfident predictions and potential misdiagnosis.Thus, these open-set samples should be treated differently, preventing unconfident predictions and potential misdiagnosis.
\section{Introduction}
Computational pathology has become a ripe ground for deep learning approaches as it has witnessed a rapid influx of myriad tasks, such as tissue phenotyping from whole slide images (WSIs). Nevertheless, even in routine clinical practice, curating huge-size WSIs with the heterogeneity of multiple tissues remains a daunting challenge.
\begin{figure}[t]
    \centering
    \includegraphics[width=\linewidth]{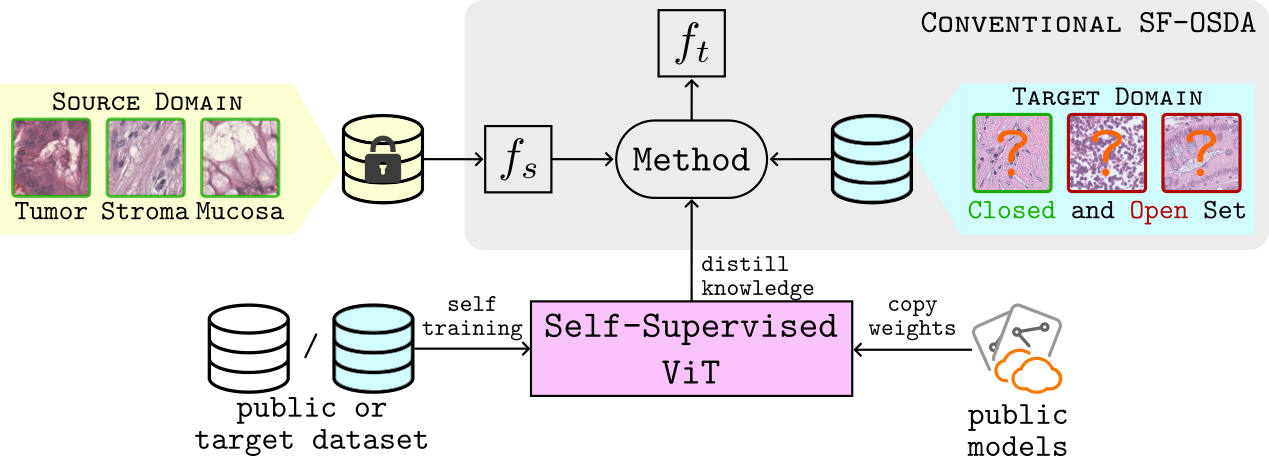}
    % \caption{\Guillaume{\textbf{Conceptual comparison of conventional SF-OSDA vs. SSL ViT-based SF-OSDA}: (a) In conventional SF-OSDA, the source model $f_s$ pre-trained on the inaccessible source domain $\mathcal{D}_s$ is adapted to the unlabeled target domain $\mathcal{D}_t$ comprising of closed-set (known) and open-set (unknown) classes; (b) SSL ViT-based SF-OSDA can avail from a self-supervised vision transformer trained on the target domain, allowing access to a stronger contextualized embedding for the target data.}}
        \caption{\textbf{Unveiling the self-supervised vision transformer (ViT) for source-free open-set domain adaptation (SF-OSDA).} The source model $f_s$ undergoes adaptation, resulting in the adapted model $f_t$ acclimating to an unlabeled target domain, accommodating both closed-set (known) and open-set (unknown) classes, all while maintaining a strict boundary of not accessing the source dataset. Our methodology capitalizes on distilling knowledge from a self-supervised ViT, leveraging its potent capability to generate contextually enriched target embeddings. Guidance for knowledge distillation can originate from two principal sources: self-supervised pre-trained transformer models without adaptation and models that have undergone extensive self-supervised pre-training on publicly available histopathology datasets or target domain data, showcasing our approach's adaptability.}
    \label{fig:1}
\end{figure}

% In this situation, transfer learning and domain adaptation techniques from open-source image datasets~\cite{kather2019predicting,javed2020cellular} can mitigate or reduce costly annotation to be focused only on the labeling of the region of interest, i.e., relevant tissues. Albeit successful, access to a huge amount of labeled open-source data may be unattainable after deployment due to regulations on data privacy and computational limitations. This heralds a practical domain adaptation scenario, namely a source-free domain adaptation (SFDA) setting, where the task is to transfer knowledge from inaccessible source data to unlabeled target data using only a source-trained model.
In this situation, transfer learning and domain adaptation techniques \cite{ganin2015unsupervised, zhang2020collaborative, hoffman2018cycada} that leverage labeled source data to adapt to unlabeled target data can help mitigate the need for expensive manual annotation. However, access to source data may be unattainable after deployment due to regulations on data privacy, storage limitations, or limitations in computational resources. Consequently, simultaneous access to both source and target domains may not be a realistic option, rendering traditional transfer learning approaches impractical.
This heralds a practical domain adaptation scenario for computational pathology, namely a source-free domain adaptation (SFDA) setting, where the task is to transfer knowledge from inaccessible source data to unlabeled target data using only a source-trained model.
Several SFDA methods~\cite{liang2020we,qu2022bmd} have been proposed en route to the goal based on test-time training/adaptation~\cite{nado2020evaluating,tomar2022opttta}.
%However, while promising, these methods follow the closed-world assumption and do not account for the open-set samples from target data that are not present in the training source set. The open-set samples are unknown novel categories outside the set of initial categories annotated by the user, which could be \textbf{abundant but clinically irrelevant tissue categories} for the task at hand from histological slides or related to rare diseases and/or caused by acquisition artifacts. Thus, these open-set samples can be manually reviewed to avoid making the model susceptible to a potential misdiagnosis.
Nonetheless, for every new dataset, these methods necessitate extensive annotation of whole slide images. Thus, it is important to note that not all tissues within these images are pertinent to the task at hand, such as abundant but clinically irrelevant tissue categories. By concentrating solely on the relevant tissues, we can significantly lighten the workload for practitioners. This suggests the need to develop a system capable of effectively filtering out open-set samples that do not belong to the initially labeled categories. Consequently, these open-set samples can be manually reviewed to ensure the model's robustness.

In addition to open-set samples, the source-trained model may also encounter a shift in color appearance (chromatic shift) between the source and target domains of histopathology images, referred to as covariate shift due to different types of scanners or staining procedures. Nevertheless, current open-set detection methods ~\cite{galdran2022test,vaze2022openset,chen2021adversarial,tack2020csi} are often incapable of disentangling \textbf{semantically shifted open-set samples}, e.g., unknown tissue categories from the \textbf{covariate shifted closed-set samples} (known tissue categories with different chromatic distribution).

To this end, we propose $\ours$, which focuses on a more practical source-free open-set domain adaptation (SF-OSDA) setting and mitigates the above challenges inherent in classifying histopathological tissue images in the presence of both covariate and semantic shifts. 
%Contrary to previous SF-OSDA methods
% (Fig.~\ref{fig:1} left),
% $\ours$ distills a self-supervised vision transformer to learn a stronger contextualized embedding for
% the target data (Fig.~\ref{fig:1}).
$\ours$ leverages a self-supervised vision transformer (ViT)~\cite{dosovitskiy2020image} to distill knowledge and generate enriched contextualized embeddings for the target data, outperforming the classifier model trained on source data. The realm of self-supervised learning has seen substantial progress with large-scale datasets, as evidenced by works such as \cite{caron2021emerging,wang2022}. This progress has enabled ViTs to emerge as robust feature extractors for novel downstream datasets, showcasing a reduced bias towards local textures compared to the convolutional neural networks (CNNs)~\cite{Naseer2021intriguing}. As shown in Fig.~\ref{fig:1}, the guidance from the self-supervised ViT can be sourced from models that have undergone rigorous pre-training on public histopathology datasets, for example~\cite{wang2022,chen2023general}. Alternatively, it can be derived from models self-trained on the target data itself employing various self-supervised learning algorithms~\cite{chen2020simple,caron2020unsupervised,caron2021emerging,oquab2023dinov2}. Our major contributions are as follows:

\begin{itemize}
    \item We propose a novel knowledge transfer technique from a self-supervised ViT to a source-trained classifier that utilizes guidance derived from either pre-trained models on public histopathology datasets or models that are self-trained on the target data. To enhance the credibility of pseudo labels generated by the source-trained classifier model, we introduce the Closed-Set Affinity Score (CSAS), utilizing the transformer’s contextualized embedding space. This allows us to generate reliable pseudo labels, facilitating the adaptation of the source model.
    % \item We propose Cluster Relative Maximum Logit Score (CRMLS) computed in the transformer's contextualized embedding space to refine the weak-pseudo labels with unreliable confidence from the source model. Consequently,  using these refined confidence scores on the target images, we obtain weighted average class prototypes in the transformer's embedding space and generate reliable pseudo labels for adapting the source model.
    %
    % \item  We propose automatic adversarial style augmentation (AdvStyle), which emulates covariate shifts resembling different chromatic staining shifts in histopathological images. The proposed data augmentation is utilized for self-supervised training of the vision transformer on the target data, yielding strongly contextualized embedding by pushing similar tissue features closer without using any labels. 
    % \item \Devavrat{We also propose automatic adversarial style augmentation (AdvStyle), emulating covariate shifts resembling different chromatic staining shifts in histopathological images. The proposed data augmentation can be utilized for self-supervised training of the vision transformer on the target data, yielding strongly contextualized embedding by pushing similar tissue features closer without using any labels.}
    \item We introduce automatic adversarial style augmentation (AdvStyle) to simulate covariate shifts resembling staining variations observed in histopathological imagery. This augmentation enhances self-supervised training of the vision transformer on target data, creating contextualized embeddings that align similar tissue features without the need for labels.
    \item Our method can be easily plugged into various open-set detection methods, consequently enhancing their performance in navigating covariate shifts within the target domain.
    \item We rigorously evaluate $\ours$ on three distinct histopathological datasets tailored for CRC assessment. The empirical findings underscore the remarkable prowess of $\ours$, as it consistently outpaces previous competing methods, encompassing open-set detection and SF-OSDA approaches. This leads to a marked improvement in performance spanning both closed-set and open-set recognition tasks, highlighting the robustness and effectiveness of our proposed solution. Our code and models will be available at \texttt{\color{magenta}{https://github.com/LTS5/Distill-SODA}}.
\end{itemize}
\section{Related Work}
%\Devavrat{This section provides a brief overview of previous works on open-set detection, Test-Time or Source-Free Domain Adaptation, and Source-Free Open-Set Domain Adaptation, where we discuss their strengths and limitations in the context of the open-world setting. Our approach is primarily aligned with Source-Free Domain Adaptation, incorporating simultaneous Open-Set recognition in the target domain.}
%
%\vspace{0.5em}
%\noindent
%\textbf{open-set Detection.}
\subsection{Open-Set Detection}\label{subsec:Open Set Detection}
In an open-set setting, a trained model must be able to discriminate known (closed-set) from unknown categories (open-set) that have not yet been encountered. There are a few subcategories of open-set recognition approaches. The first is model activation rectification strategies \cite{bendale2016towards, sun2021react}, where open-set examples can be detected using a rectification scheme for the activation patterns of the model related to closed-set examples. Other approaches utilize GANs to generate images resembling open-set samples \cite{neal2018open} or train GAN-discriminator to distinguish closed from open-set samples \cite{kong2021opengan}. Seminal work \cite{vaze2022openset} demonstrates that robust classifiers with high closed-set accuracy also serve as better open-set detectors. The authors establish Maximum Logit Score (MLS) as a baseline open-set scoring metric for distinguishing known from unknown classes. Motivated by this, our method uses un-normalized logits for model adaptation and open-set recognition instead of the softmax output probabilities.

Other methods, such as Monte Carlo Dropout or Multi-head CNN-based models \cite{devries2018learning, linmans2020efficient}, measure uncertainty during inference to recognize open-set images with high uncertainty. Another subcategory of methods evaluates the model on pretext tasks, e.g., predicting geometric \cite{golan2018deep} or color transformations \cite{galdran2022test} applied to test images during inference, with the assumption that the model will perform poorly on open-set samples. Nonetheless, most of these methods require a specific source model training or access to closed-set examples, limiting their applicability with off-the-shelf pre-trained models. Additionally, these methods assume similar image characteristics in the test domain, leading to failure under covariate shifts.

%\vspace{0.5em}
%\noindent
%\textbf{Source-Free Test-Time  Domain Adaptation.}
% \subsection{Source-Free Test-Time  Domain Adaptation}
% Test-time domain adaptation, known as source-free domain adaptation (SFDA) \cite{kundu2020universal,chen2022contrastive, liang2020we}, aims to improve model robustness against distribution shifts during inference without access to labeled source domain images. While current pseudo-label-based SFDA methods \cite{chen2022contrastive, liang2020we, tomar2023tesla, wang2022continual, yuan2023robust} work well in closed-world settings, they fail significantly in open-world scenarios and require filtering of open-set examples during inference.

% Moreover, another subcategory of SFDA methods, e.g., \cite{sun2020test, liu2021ttt++, su2022revisiting}, solve particular auxiliary tasks (e.g., self-supervised rotation prediction) to adapt the model under common distributional shifts. Some other methods adapt batch normalization (BN) statistics \cite{nado2020evaluating} or proposed test-time augmentation to simulate augmentations resembling the saved BN statistics \cite{ tomar2022opttta} to mitigate errors from open-set examples. However, as shown in our experiments, solely bridging the target-source feature statistics gap may not be sufficient for improving open-set recognition capabilities.

%\vspace{0.5em}
%\noindent
%\textbf{Source-Free Open Set Domain Adaptation.}
\subsection{Source-Free Open-Set Domain Adaptation}
Existing SF-OSDA methods often use \rev{prototypical-based pseudo-labeling \cite{liang2020we, zhang2021prototypical, lin2023prototypical}},  uncertainty quantification in the source model prediction \cite{roy2022uncertainty}, or proposed specialized source training strategy \cite{kundu2020towards} to train an inheritable model capable of adapting to the target domain with novel categories. 
\rev{As a seminal work, SHOT \cite{liang2020we} introduced an SF-OSDA method that employs a self-supervised pseudo-labeling scheme that clusters known and unknown categories using 2-means clustering, which is based on the entropy of predictions.} Subsequently, the model is adapted exclusively to examples from the known categories in the target domain. \rev{In \cite{zhang2023class}, inter-class relationships are modeled by leveraging the classifier layer weights of the source model, and this information is combined with contrastive learning to pseudo-label target domain images for adaptation. Recent SF-OSDA advancements, such as the Global and Local Clustering (GLC) method \cite{qu2023upcycling}, rely on carefully tuned thresholds used as reference and training hyperparameters. However, this heavy reliance on thresholds may impede the model's adaptability and reliability when dealing with diverse and unpredictable new data categories. Another recent SF-OSDA method is the source-free progressive graph learning method \cite{luo2023source}. This method incorporates a Graph Neural Network (GNN) to label images in the target domain sequentially. Nevertheless, its resource-intensive nature, involving multiple episodes of populating target samples for known and unknown classes, poses challenges.}

\rev{A common major limitation in the abovementioned methods is their reliance on the source model's feature representation to rectify the target dataset's pseudo-labels. Since the source model's feature representation on the covariate-shifted target domain is poor, the performance gain against the baseline source model may be only marginal. In contrast to these limitations, our approach distinguishes itself by leveraging an external self-supervised vision transformer to create refined pseudo labels through semantic clustering without ground truth labels. This not only improves the differentiation between open-set and closed-set images but also enables the application of our method to any pre-trained source model, enhancing the performance of various open-set detection techniques.}

\section{Materials and Methods}
% \Devavrat{Adapting a source model pre-trained on the inaccessible source data to the unlabeled target data of histopathological images under simultaneous \textit{covariate} and \textit{semantic} shifts is extremely challenging as the uncertainty of the source model's predictions on the target domain's images may come from either or both types of distributional shifts. Overcoming this challenge, we propose distilling knowledge from a vision transformer (ViT)~\cite{dosovitskiy2020image} based feature extractor capable of generating strongly contextualized embedding in the target domain. Recent developments in the field of self-supervised learning on large-scale datasets, e.g.,~\cite{caron2021emerging, wang2022}, have yielded vision transformers that can act as strong feature extractors for unseen downstream datasets, demonstrating better robustness against, e.g., texture bias than CNN-based models~\cite{Naseer2021intriguing}. Such vision transformer-based feature extractors may be obtained from public repositories~\cite{wang2022, chen2023general}, or fine-tuned on the target / public histopathology datasets using self-supervised learning algorithms~\cite{chen2020simple,caron2020unsupervised,caron2021emerging, oquab2023dinov2}}
\begin{figure*}[!t]
    \centering
    \includegraphics[width=0.8\linewidth]{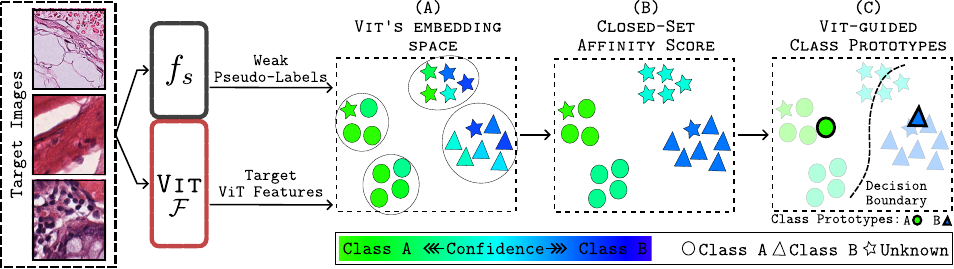}
    \caption{\textbf{ViT guided closed-set class prototypes.} Utilizing the self-supervised ViT feature extractor $\mathcal{F}$ in conjunction with the source model $f_s$, we first obtain the contextualized embeddings for target images via $\mathcal{F}$, which are weakly labeled by $f_s$ and subsequently grouped into $K$ clusters using $K$-means clustering. Leveraging the closed-set affinity score (CSAS) detailed in Section \ref{subsec:csas}, we refine the confidence of weak pseudo labels and compute weighted class prototypes of closed-set (known) classes within the embedding space of  $\mathcal{F}$. Note that the final closed-set class prototypes exhibit improved decision boundaries, with samples from open-set classes being assigned low confidence scores. In this illustration, the three shapes represent closed-set classes A, B, and open-set class, with the color intensity signifying the source model's confidence level in categorizing the target images into known classes A and B.}
    \label{fig:crmls} 
\end{figure*}
\noindent
\textbf{Problem formulation.} Let $f_s: X\to Y$ denote the tissue-type classifier (source model) 
 that is trained on the inaccessible source domain $\mathcal{D}_s$ of tissue images belonging to $C$ known classes, where $X\in \mathbb{R}^{H\times W\times 3}$ denotes an RGB image with height $H$ and width $W$, and $Y\in \mathbb{R}^{C}$ denotes the corresponding logits vector. Also, let $\mathcal{D}_t$ denote the unlabeled target domain images belonging to the $C$ known and $\Bar{C}$ unknown classes not seen during source model training, i.e., $C\cup \Bar{C}$. We aim to adapt pre-trained source model $f_s$ to the target domain $\mathcal{D}_t$ to obtain adapted model $f_t$ so that it can correctly classify target domain images into $C$ known classes while also identifying images that fall into an open-set category comprising $\Bar{C}$ unknown classes. 

\subsection{Closed-Set Class Prototypes in ViT Embedding Space}
In the following, we assume access to a strong ViT-based feature extractor $\mathcal{F}: X\to Z$ pre-trained in a self-supervised manner on the target domain $\mathcal{D}_t$ or public histopathology dataset $\mathcal{D}_p$. Here $Z\in \mathbb{R}^D$ denotes the ViT feature space of dimension $D$. In addition, we re-calibrate the batch normalization \cite{ioffe2015batch} statistics of the source model $f_s$ with that of the target domain if $f_s$ is a standard CNN architecture. Re-calibrating batch normalization statistics of $f_s$ can effectively reduce the feature distribution shifts encountered in CNNs \cite{nado2020evaluating}.

We begin by projecting the target images into the ViT's embedding space, $\mathcal{F}$ \rev{(see Fig. \ref{fig:crmls}A)}. Our objective here is to compute class prototypes within this space, utilizing the predictions provided by the source model. The class prototypes are represented as $P = [p_1, p_2, \dots, p_C]$, and are derived through the following:
\begin{equation}\label{eq:class_prototypes_gen}
    p_j = \frac{\sum_{x \in \mathcal{D}_t} \mathbf{1}[\arg\max f_s(x) = j] \cdot \zeta (x) \cdot \mathcal{F}(x)}{\sum_{x \in \mathcal{D}_t}\mathbf{1}[\arg\max f_s(x) = j] \cdot \zeta(x)}
\end{equation} 
where $p_j$ represents the prototype of closed-set class $j$, and $\zeta(x)\geq 0$ serves as the weight function utilized in computing the \textit{average}. Depending on the choice of $\zeta$, we can derive different variations for calculating the closed-set class prototypes:
\begin{enumerate}
    \item \textbf{Uniform Weighting.} Setting $\zeta(x) = 1$ for all $x\in \mathcal{D}_t$, we derive the closed-set class prototypes by averaging the target ViT features that correspond to the particular class, as determined by the source model $f_s$.
    
    \item \textbf{Maximum Softmax Probability Weighting.} In this scenario, $\zeta(x)$ is defined as $\max \mathop{softmax}(f_s(x))$. Rather than applying equal weight to the target features in prototype computation, this method assigns weights based on the probability that each feature belongs to the closed-set class, according to predictions made by the source model $f_s$.
    
    \item \textbf{Positive Maximum Logit Score Weighting.} Given that the unnormalized maximum raw logit output from $f_s$ has demonstrated efficacy in identifying open-set examples~\cite{vaze2022openset}, we set $\zeta(x) = \max f_s(x) - \min_{x \in \mathcal{D}_t} \max f_s(x)$ for this weighting scheme. Note that we ensure $\zeta(x)\geq0$ as maximum logit score can be negative.
\end{enumerate}

One major limitation when employing any of the previously mentioned formulations under covariate shift conditions is the introduction of noise and uncertainty in the confidence of the classes assigned by $f_s$ to target images. While we operate under the assumption that semantically similar examples naturally group together in the ViT $\mathcal{F}$'s embedding space, the presence of covariate shift can cause $f_s$ to inconsistently label semantically alike examples within this space. Consequently, this inconsistency can lead to the inaccurate calculation of closed-set class prototypes, mistakenly incorporating features from open-set targets. To address and rectify this issue, we suggest the following three-step approach:

\subsubsection{Enhancing Semantic Consistency in ViT Embedding Space via K-means Clustering}
Building on our previous discussion, the self-supervised ViT assigns similar feature representations to the target images that share semantic characteristics. We leverage this trait of the ViT embedding space to refine the predictions made by the source model $f_s$ in the target domain. To do this, we first partition all target images $\mathcal{D}_t$ into $K$ disjoint clusters as $\{\mathcal{D}_1, ..., \mathcal{D}_K\}$ via the $K$-means clustering algorithm, and subsequently represent each target image based on the attributes of its respective cluster.

\subsubsection{Defining Cluster Attributes: Closed-Set Affinity Score (CSAS), Class Prior, and Class Conditional Mean}\label{subsec:csas}
For each cluster $\mathcal{D}_k$, we define three key attributes that will aid in generating improved closed-set class prototypes.
\paragraph{{Closed-Set Affinity Score (CSAS)}} The Closed-Set Affinity Score $\Phi_\text{CSAS}$ for a cluster $\mathcal{D}_k$ is defined as follows:
\begin{equation}\label{eq:csas}
     \Phi_\text{CSAS}(\mathcal{D}_k) \leftarrow \max {\mathbb{E}}_{x\sim \mathcal{D}_k} \{f_s(x)\}
\end{equation} 
where $\max$ is taken over $C$ closed-set classes. A high $\Phi_\text{CSAS}$ score for a cluster indicates that the target images within it consistently receive similar predictions from the source model $f_s$, suggesting a strong affiliation with one of the closed-set classes. Conversely, a low score implies that the cluster comprises target images with diverse and/or average low confidence predictions, hinting at a potential open-set class association\rev{, as illustrated in Fig. \ref{fig:crmls}B.}

\paragraph{{Class Prior}} The closed-set class prior for class $j$ in cluster $\mathcal{D}_k$ is calculated as the ratio of target images in $\mathcal{D}_k$ that are classified as class $j$ by the source model $f_s$.
\begin{equation}
    \alpha_{\text{prior}}(\mathcal{D}_k, j) \leftarrow \frac{\sum_{x \in \mathcal{D}_k} \mathbf{1}[\arg\max f_s(x) = j]}{|\mathcal{D}_k|}
\end{equation}
\paragraph{{Class Conditional Mean}} The class conditional mean for each cluster is computed as the mean target feature of the closed-set classes belonging to the same cluster, as predicted by the source model $f_s$.
\begin{equation}
    \mu(\mathcal{D}_k, j) \leftarrow \frac{\sum_{x \in \mathcal{D}_k} \mathbf{1}[\arg\max f_s(x) = j] \cdot \mathcal{F}(x)}{\sum_{x \in \mathcal{D}_k} \mathbf{1}[\arg\max f_s(x) = j]}
\end{equation}

\subsubsection{Computing closed-set class prototypes using cluster attributes} Utilizing the trio of attributes defined for each cluster within the ViT embedding space, we have the capability to calculate the prototypes for the closed-set classes. This process incorporates the consideration of each cluster's influence on all the closed-set classes and also discerns whether a cluster is associated with an open-set class. The prototype for a class $j$ is defined as:
\begin{equation}\label{eq:class_proto_csas}
    p_j = \frac{\sum_{k=1}^{K}\alpha_{\text{prior}}(\mathcal{D}_k, j) \cdot \Hat{\Phi}_\text{CSAS}(\mathcal{D}_k) \cdot \mu(\mathcal{D}_k, j)} { \sum_{k=1}^{K}\alpha_{\text{prior}}(\mathcal{D}_k, j) \cdot \Hat{\Phi}_\text{CSAS}(\mathcal{D}_k) }
\end{equation} 
where $\Hat{\Phi}_\text{CSAS}(\mathcal{D}_k)$ represents the normalized Closed-Set Affinity Score for cluster $\mathcal{D}_k$, calculated as:
\begin{equation}\label{eq:normalized_csas}
    \Hat{\Phi}_\text{CSAS}(\mathcal{D}_k)=\Phi_\text{CSAS}(\mathcal{D}_k) - |\mathcal{D}_k|\cdot \min_i \frac{\Phi_\text{CSAS}(\mathcal{D}_i)}{|\mathcal{D}_i|} 
\end{equation} 
where $|\mathcal{D}|$ denotes size of the set $\mathcal{D}$.
Note that the above expression can be reformulated in the subsequent manner:
\begin{align}
    p_j &= \frac{\sum_{k=1}^{K} \frac{\sum_{x \in \mathcal{D}_k} \mathbf{1}[\arg\max f_s(x) = j] \cdot \mathcal{F}(x)}{|\mathcal{D}_k|} \cdot \Hat{\Phi}_\text{CSAS}(\mathcal{D}_k)} { \sum_{k=1}^{K} \frac{\sum_{x \in \mathcal{D}_k} \mathbf{1}[\arg\max f_s(x) = j]}{|\mathcal{D}_k|} \cdot \Hat{\Phi}_\text{CSAS}(\mathcal{D}_k)} \nonumber \\
    &= \frac{\sum_{x \in \mathcal{D}_t} \sum_{k=1}^{K} \frac{\mathbf{1}[x\in \mathcal{D}_k] \cdot \mathbf{1}[\arg\max f_s(x) = j] \cdot \mathcal{F}(x) \cdot \Hat{\Phi}_\text{CSAS}(\mathcal{D}_k)}{|\mathcal{D}_k|}}{\sum_{x \in \mathcal{D}_t} \sum_{k=1}^{K} \frac{\mathbf{1}[x\in \mathcal{D}_k] \cdot \mathbf{1}[\arg\max f_s(x) = j] \cdot \Hat{\Phi}_\text{CSAS}(\mathcal{D}_k)}{|\mathcal{D}_k|}} 
\end{align} 
%
% By comparing this to Eq. \ref{eq:class_prototypes_gen}, we determine $\zeta(x)$ to be:
\rev{This enables the computation of class prototypes solely based on CSAS (see Fig. \ref{fig:crmls}C) by defining the $\zeta$ in Eq. \ref{eq:class_prototypes_gen} as follows:}
\begin{equation}\label{eq:csas_zeta}
    \zeta(x)= \sum_{k=1}^{K} \mathbf{1}[x\in \mathcal{D}_k] \frac{\Hat{\Phi}_\text{CSAS}(\mathcal{D}_k)}{|\mathcal{D}_k|}
\end{equation} 

Note that $\zeta(x) \geq 0 \hspace{1em} \forall x \in \mathcal{D}_t$. This holds true as the ratio $\frac{\Hat{\Phi}_\text{CSAS}(\mathcal{D}_k)}{|\mathcal{D}_k|}\geq0$ derived from Eq. \ref{eq:normalized_csas}.

\subsection{Source-Free Adaptation via ViT Guided Closed-Set Class Prototypes}\label{subsec:sfosda}
% To obtain the pseudo-labels for self-training the source model $f_s$ on the target domain images, we first compute the class prototypes $P = [p_1, p_2, \dots p_C]$ of $C$ known classes in the transformer's contextualized embedding space using CRMLS scores. Let $X^j = \{\forall x|x\in \mathcal{D}_t \land \arg\max f_s(x) = j\}$ denote all target domain images that are classified by $f_s$ as class $j$. Also, $\forall x \in X^j$, let $\Phi_\text{CRMLS}(x)$ denote its confidence score and $S_\text{cluster}(x) = |\mathcal{D}_{i \ni x \in \mathcal{D}_i}|$ denote the size of the cluster containing $x$. We compute the prototype of class $j$ as:
%
% 
% \begin{equation}\label{eq:class_proto}
%     p_j = \frac{\sum_{x \in X^j}\Hat{w}(x)\cdot \mathcal{F}(x)}{\sum_{x \in X^j} \Hat{w}(x)}
% \end{equation}
% where,
% \begin{equation*}
%     \Hat{w}(x) = w(x) - \mathop{\min}_{x \in \mathcal{D}_t} w(x), \hspace{2em}
%     w(x)=\frac{\Phi_\text{CRMLS}(x)}{S_\text{cluster}(x)}
% \end{equation*}

% %
Due to instability caused by $K$-means clustering initialization, we run $N_\text{mc}$ Monte-Carlo simulations of $K$-means clustering (see ablations in Fig.~\ref{fig:ablation_csas} (b)-(c)). This process is employed to compute $\Phi_\text{CSAS}(\mathcal{D}_k)$ as per Eq. \ref{eq:class_proto_csas} and CSAS-based $\zeta(x)$ as indicated in Eq. \ref{eq:csas_zeta}, thereby achieving more accurate estimates for the target domain images $x\in \mathcal{D}_t$. 
% $\Hat{w}(x)$ ensures the weights are positive and dividing $\Phi_\text{CRCLS}(x)$ by the corresponding cluster size $S_\text{cluster}(x)$ ensures the prototypes are not biased towards large clusters.

Let $P_\text{norm}=[\hat{p}_1, \hat{p}_2, \dots \hat{p}_C]$ denote unit norm closed-set class prototypes such that $\hat{p}_j=p_j/\|p_j\|$. We compute the un-normalized pseudo-logits $\Bar{y}$ for the target image $x$ as follows:
\begin{equation}\label{eq:proto_logits}
\Bar{y}(x) = P_\text{norm}^{T}\frac{\mathcal{F}(x)}{\|\mathcal{F}(x)\|_2} / \tau 
\end{equation}
where $\tau$ is a temperature hyperparameter. The source model $f_s$ is adapted on the target domain images $\mathcal{D}_t$ using the pseudo-logits of Eq. \ref{eq:proto_logits} to obtain $f_t^*$ by minimizing the following mean squared error (MSE) loss:
\begin{equation}\label{eq:self_distill}
    f_t^* = \argmin_f \mathbb{E}_{x \sim \mathcal{D}_t} \|f(x) - \Bar{y}(x)\|_2^2
\end{equation} 
We opt for MSE loss instead of the KL-divergence loss as the former allows better open-set detection based on \textit{Maximum Logit Score} (see ablation in Table \ref{tab:loss}).

\subsection{Self-Supervised ViT Training via Automatic Adversarial Style Augmentation
}\label{subsec:ssl_adv_style}
We adopt DINO \cite{caron2021emerging} for self-supervised training of our feature extractor $\mathcal{F}$ on the unlabeled target or public histopathology dataset. The choice of DINO is due to its effectiveness as a nearest neighbor classifier, adept at creating clusters of semantically similar images within its embedding space. DINO uses an identical architecture of teacher-student networks, where the teacher network is slowly updated as a moving average by the student network during training. The soft-maxed logits predicted by the teacher network on randomly augmented global crops of the image views are matched by the student network on another set of augmented image views with global and local image crops. 

In our approach, we enhance the DINO framework by substituting its standard augmentations with our custom-designed adversarial style augmentation (AdvStyle), which is tailored for the extraction of contextually rich embeddings from target images. We maintain the original DINO hyper-parameters for fine-tuning the vision transformer on either the target or the public dataset, ensuring consistency and stability in the training process.

\vspace{0.5em}
\noindent
\textbf{Automatic adversarial style augmentation (AdvStyle).} Using the data augmentation policies from \cite{galdran2022test}, let $\mathbb{O}$ represent the set of color transformations operations $\mathcal{O}: X \to X$. We define $\mathbb{O}$ as the set of six color transformations, including $\left \{ \footnotesize{ \textsc{Gamma, Hue, Saturation, Sharpness, Brightness, Contrast}} \right \}$. Each color transformation in $\mathbb{O}$ is applied to a given tissue image parametrized by its magnitude $\Hat{m}\in \left [ 0,1 \right ]$, which determines the strength of the color transformation. We utilize a style augmentation module consisting of a shallow multilayer perception (MLP) $f_\text{style}$ built on top of the frozen ViT teacher network that sequentially applies differentiable image transformation operations $\mathcal{T}(x, \hat{m})$ to input image $x$ using their predicted magnitudes $\hat{m}$ as follows:
\begin{equation}
    \Hat{x} = \mathcal{T}(x, \Hat{m}) = \mathcal{O}_6(\mathcal{O}_5 ...(\mathcal{O}_1(x, \Hat{m}_1), ..., \Hat{m}_5, \Hat{m}_6)
\end{equation}
where $\mathcal{O}_j(x, \Hat{m}_j)$ applies $j^\text{th}$ color transform with learnable magnitude $\Hat{m}_j$ on a given target domain image $x$ to generate augmented image $\Hat{x}$.

\begin{figure}[t]
    \centering
    \includegraphics[width=0.75\linewidth]{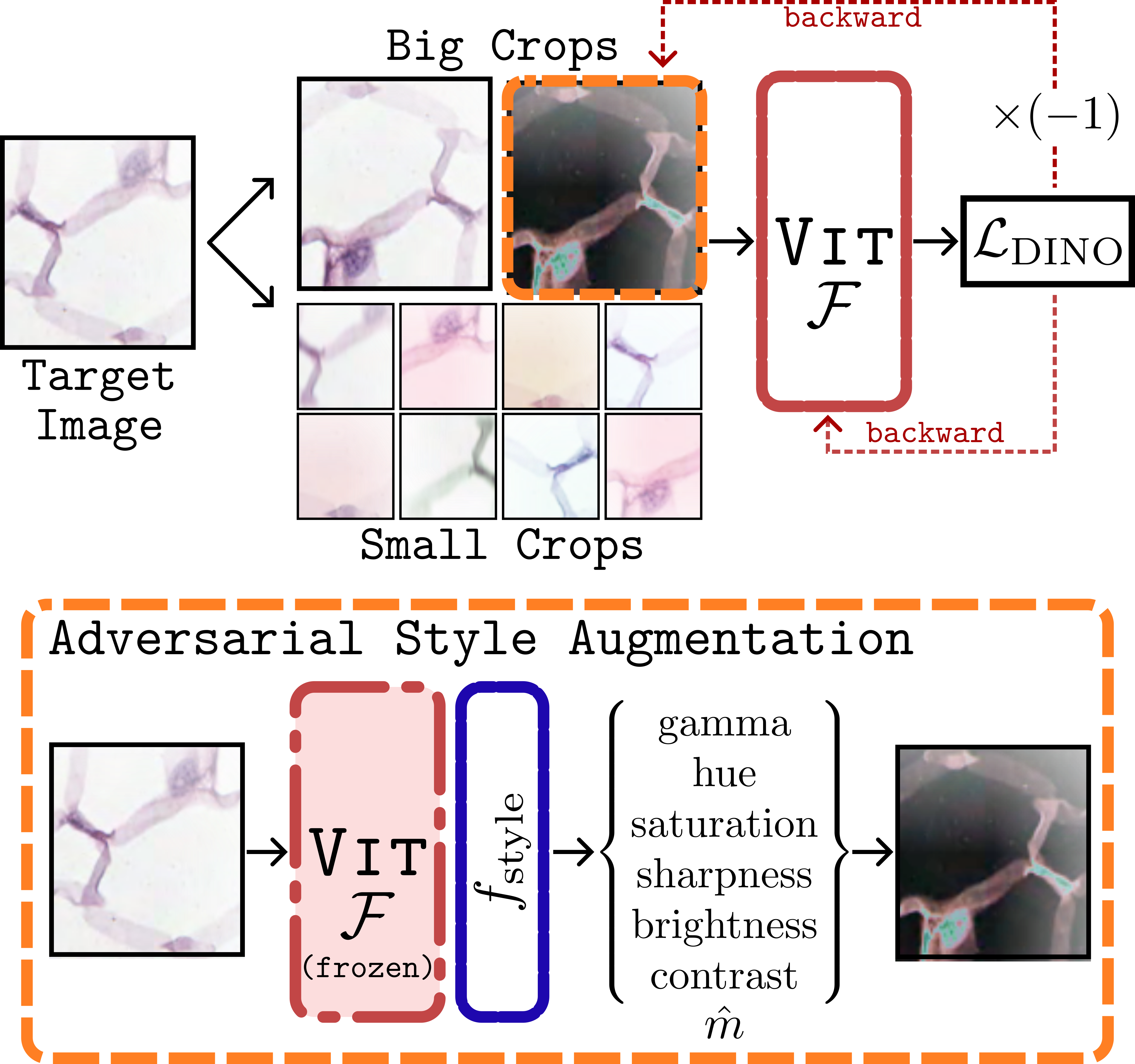}
    \caption{\textbf{Vision transformer self-training with adversarial style augmentations.} The style augmentation module $f_\text{style}$ is trained to learn magnitudes $\Hat{m}$ for adversarial augmentations by maximizing $\mathcal{L}_\text{DINO}$. Concurrently, the ViT encoder  $\mathcal{F}$ is updated to minimize $\mathcal{L}_\text{DINO}$ on the target domain, establishing a dynamic adversarial training setup.}
    \label{fig:vit_train}
\end{figure}

As shown in Fig. \ref{fig:vit_train}, we self-train ViT encoder $\mathcal{F}$ on $\mathcal{D}_t$ using the same self-supervised DINO loss $\mathcal{L}_\text{DINO}$ (Eq. \ref{eq:dino_loss}) as \cite{caron2021emerging} but with our AdvStyle instead of their default augmentations. $f_\text{style}$ is trained in an adversarial manner w.r.t. ViT encoder to generate hard style-based augmentation policies that make distinguishing augmented image pairs in the feature space difficult. In this adversarial setup, $f_\text{style}$ is trained to maximize $\mathcal{L}_\text{DINO}$ (Eq. \ref{eq:dino_loss}), while the ViT model is trained to minimize $\mathcal{L}_\text{DINO}$. Note that we only learn to adversarially augment one of the two global image crops while randomly augmenting the other global and local crops using the default color jittering as shown in Fig. \ref{fig:vit_train}. To maintain the integrity of the training process and prevent the DINO algorithm from exclusively adapting to unrealistic augmentation styles, we strategically apply adversarial augmentation to just one of the global views rather than both. This approach ensures that DINO continues learning from a balanced representation that includes original image features and augmented styles, \rev{thus avoiding model collapse.} %\Devavrat{We adversarially augment only one global view instead of both global views to ensure DINO does not cheat and learn to embed only unrealistic augmentation styles.} 
$\mathcal{L}_\text{DINO}$ is defined as follows:
\begin{equation}\label{eq:dino_loss}
\begin{split}
\mathcal{L}_\text{DINO}(x) =\hspace{0.5\linewidth}\\
\sum_{i=1}^2 \Big(\sum_{\substack{j=1 \\ j\neq i}}^2 H(G_t(x_g^i), G_s(x_g^j)) + \sum_{k=1}^{8}
H(G_t(x_g^i), G_s(x_l^k))\Big)
\end{split}
\end{equation}
where,
\begin{equation*}
\begin{split}
    x_{g}^1 = \text{Crop}_g(\text{aug}(x)), &\quad x_{g}^2 = \mathcal{T}(x, f_\text{style}(\mathcal{F}_t(x)), \\
    &\quad x_{l}^k = \text{Crop}_l(\text{aug}(x))
\end{split}
\end{equation*}
Here, $G_t$ and $G_s$ denote the teacher and student networks, each composed of a feature extractor $\mathcal{F}$ and a projection head $g: Z \to U$ applied on $\mathcal{F}$'s output $Z$, $U\in\mathbb{R}^{d'}$. $\text{Crop}_g(.)$ and $\text{Crop}_l(.)$ denote random global and local crops, while $\text{aug}(.)$ denotes random color jittering transformation applied to the target domain's images with a probability of $0.5$. $H(p,q)=-\sum p\log q$ denotes the cross-entropy loss. 
%
%\Devavrat{The learned transformations are applied sequentially on the unlabeled target domain image $x$}.

The parameters of the student network $\theta_{G_s}$ and that of adversarial augmentation module $\theta_{f_\text{style}}$ are updated to optimize the following mini-max objective:
\begin{equation}
\min_{\theta_{G_s}}\max_{\theta_{f_\text{style}}}\mathbb{E}_{x \sim \mathcal{D}_t} \big[\mathcal{L}_\text{DINO}(x)\big]
\end{equation}
The teacher network's parameters $\theta_{G_t}$ are updated by that of the student network $\theta_{G_s}$ with momentum $\nu=0.996$ after every gradient step as follows:
\begin{equation}
    \theta_{G_t} = \nu \cdot \theta_{G_t} + (1 - \nu) \cdot \theta_{G_s}
\end{equation}
%
%\section{Experimental Setup and Results}
\begin{table*}[t]
    \centering
    \caption{\textbf{Summary of datasets repartitions.} For all the datasets used in our experiments, we report the class ratio and number of image patches in train, validation, and test sets of every closed-set and open-set split -- Splits 1, 2, and 3. \\\color{gray}$^\ast$ The open-set validation subset is not used in our experiments.}  
    %For all the datasets used in our experiments, we report the number of image patches in train, validation, and test subsets of every Closed and open-set splits -- Split 1, 2, and 3.} \\\color{blue}$^\ast$Validation open-set is never used in our experiments as Validation Set is only necessary for source model training.
    \label{tab:splits}
    \resizebox{0.95\textwidth}{!}{
    \begin{tabular}{cc|c|ccc|c|ccc|c|ccc}
    \toprule
    & & \multicolumn{4}{|c}{Kather-16} & \multicolumn{4}{|c}{Kather-19} &\multicolumn{4}{|c}{CRCTP} \\
    \cmidrule{3-14}
         & & & Train & Validation & Test & & Train & Validation & Test & & Train & Validation & Test \\
    \cmidrule{4-6}\cmidrule{8-10}\cmidrule{12-14}
         & & Ratio & 70\% & 15\% & 15\% & Ratio & 70\% & 15\% & 15\% & Ratio & 70\% & 15\% & 15\% \\
    \midrule
    \multirow{2}{*}{Split 1} 
     & closed-set & 50.0\% & 1,756 & 372 & 372 & 58.6\% &41,039 & 8,790 & 8,790 & 92.9\% & 127,400 & 27,300 & 27,300 \\
     & open-set & 50.0\% & 1,756 & \color{gray}372$^\ast$ & 372 & 41.4\% & 28,971 & \color{gray}6,205$^\ast$ & 6,205 & 7.1\% & 9,800 & \color{gray}2,100$^\ast$ & 2,100 \\
     \midrule
     \multirow{2}{*}{Split 2} 
     & closed-set & 25.0\% & 878 & 186 & 186 & 38.3\% & 26,813 & 5,743 & 5,743 & 71.4\% & 98,000 & 21,000 & 21,000\\
     & open-set & 75.0\% & 2,634 & \color{gray}558$^\ast$ & 558 & 61.7\% & 43,197 & \color{gray}9,252$^\ast$ & 9,252 & 28.6\% & 39,200 & \color{gray}8,400$^\ast$ & 8,400 \\
     \midrule
     \multirow{2}{*}{Split 3} 
     & closed-set & 37.5\% & 1,317 & 279 & 279 & 49.9\% & 34,904 & 7,476 & 7,476 & 82.1\% & 112,700 & 24,150 & 24,150 \\
     & open-set & 62.5\% & 2,195 & \color{gray}465$^\ast$ & 465 & 50.1\% & 35,106 & \color{gray}7,519$^\ast$ & 7,519 & 17.9\% & 24,500 & \color{gray}5,250$^\ast$ & 5,250 \\
    \bottomrule
    \end{tabular}}
\end{table*}
\section{Experiments and Results}
\subsection{Datasets}

\begin{figure*}
    \centering
    \includegraphics[width=0.8\linewidth]{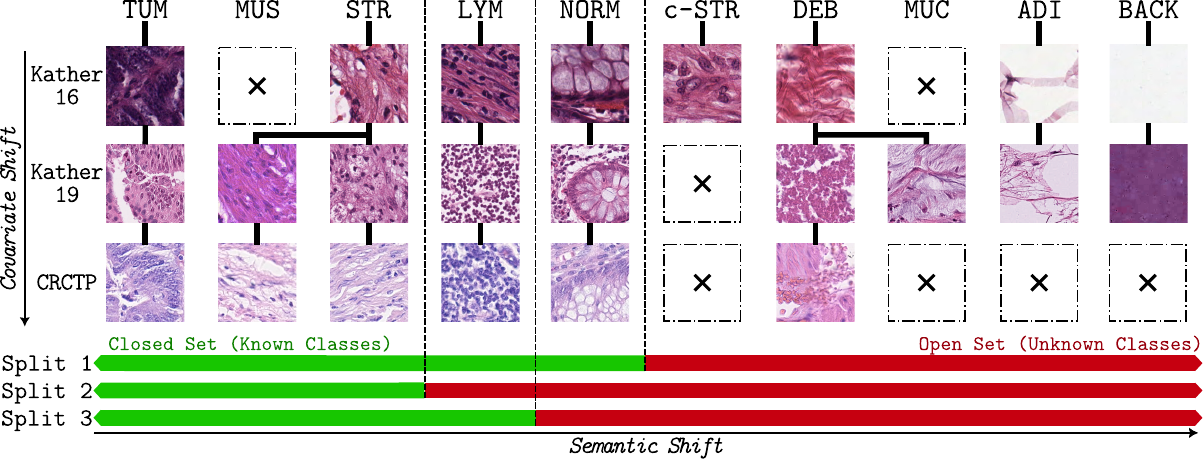}
    \caption{\textbf{Breakdown of dataset splits:} This illustrates how the dataset was segmented for the purposes of closed-set (indicated in \textcolor{green}{green}) and open-set (shown in \textcolor{red}{red}) within the three CRC datasets for tissue-type classification.}
    \label{fig:dataset}
\end{figure*}

We evaluate $\ours$ on the task of CRC tissue phenotyping using Hematoxylin and Eosin (H\&E) stained tissue sections extracted from WSIs of colorectal biopsies. 
In particular, we utilize three publicly available CRC tissue characterization datasets digitized at a magnification of $20\times$: Kather-16 \cite{kather2016multi}, Kather-19 \cite{kather2019predicting}, and CRCTP \cite{javed2020cellular}. These specific datasets were chosen as they effectively represent diverse real-world domain shifts, providing a robust testing ground for the classification of various CRC tissue types. The Kather-16 dataset comprises 5,000 patches (150$\times$150 pixels, 0.495 µm/pixel) representing eight tissue classes, with a balanced distribution of 625 patches per class. Kather-19 consists of 100,000 tissue patches (224$\times$224 pixels, 0.5 µm/pixel) divided almost evenly among nine classes. CRCTP contains 196,000 image patches (150$\times$150, 0.495 µm/pixel) categorized into seven tissue phenotypes. For the experiments, we apply a random stratified split to divide all datasets into training (70\%), validation (15\%), and test (15\%) sets. Due to discrepancies in class definitions, following consultation with expert pathologists, we follow the same harmonization approach in \cite{ABBET2022102473}, resulting in a set of 7 common classes: tumor epithelium (TUM), stroma (STR), lymphocytes (LYM), normal colon mucosa (NORM), complex stroma (c-STR), debris (DEB), and background (BACK). To make correspondence between datasets, following \cite{ABBET2022102473}, we merge stroma and smooth muscle (MUS) classes as stroma (STR) and debris and mucus (MUC) as debris (DEB). As depicted in Fig. \ref{fig:dataset}, there are notable disparities in the visual appearances of tissues across the three datasets, which can likely be attributed to variations in the staining procedure, variability in tissue preparation, or other potential factors such as the presence of folded tissues \cite{ABBET2022102473}.
\vspace{0.5em}
\noindent
\textbf{Closed-Set and Open-Set splits.} 
We adopt the experimental setup outlined in \cite{galdran2022test}
to define closed-set and open-set splits for each dataset. The three splits depicted in Fig. \ref{fig:dataset} follow the same principles described in \cite{galdran2022test}, with minor modifications. Specifically, {Split 1} emulates a scenario where a practitioner labels only clinically relevant tissue regions (TUM, STR, LYM, and NORM) used for the critical task of, e.g., quantifying tumor-stroma ratio \cite{van2018tumour, abbet2022toward}, leaving uninformative regions (c-STR, DEB, ADI, BACK) unlabeled. Notably, in contrast to \cite{galdran2022test}, c-STR is considered an open-set category because it is not present in Kather-19 and CRCTP datasets \cite{ABBET2022102473}. {Split 2} and {Split 3} complement the analysis by focusing on the classification of tumoral regions (TUM and STR) while excluding healthy tissues (LYM, MUC) and uninformative but abundant samples (c-STR, DEB, ADI, BACK). Additionally, {Split 3} includes lymphocyte images (LYM) in the closed-set of {Split 2} to introduce more challenging closed-set classification scenarios. Table \ref{tab:splits} reports each dataset repartition and size, along with the corresponding Closed and open-set class ratios. The open-set class ratio ranges from 7.1\% to 75\%, allowing us to cover various SF-OSDA scenarios.

\subsection{Evaluation Protocol}
% {\textbf{Evaluation Protocols.}} We utilize two evaluation metrics: class Average Closed-Set Accuracy (ACC) for assessing domain adaptation competency and area under RoC Curve (AUC) for detecting open-set categories. Using each dataset repartition for the respective Open/closed-set splits, the source model $f_S$ is trained on the source training set of only closed-set classes, and the best-performing model on the source validation set is chosen for adaptation. $f_s$ is then adapted on the unlabeled target domain training set (containing both closed-set and open-set) and evaluated on the target domain test set. Results are averaged over 10 random seeds of source model training.

We evaluate all baselines and our proposed $\ours$ \rev{with the following protocol.} Firstly, we train the source model $f_s$ on the training closed-set and validate it on the validation closed-set of the source domain for selecting the best-performing source model. Next, we adapt $f_s\to f_t$ on the target domain's training subset, including closed-sets and open-sets. Finally, we evaluate the performance of $f_t$ on the test subset of the target domain. We use class average closed-set accuracy (ACC) to assess domain adaptation on the closed-set and area under the ROC curve (AUC) to access open-set detection. We perform six adaptation experiments, including pairwise adaptation among Kather-16, Kather-19, and CRCTP datasets on every split. We report results averaged over ten random seeds of source model training.

\subsection{Baselines}
We compare our method against several seminal and state-of-the-art (SOTA) methods for open-set detection (OSR) and SF-OSDA. Unlike SF-OSDA strategies, OSR methods are solely tested on the target domain without any adaptation. As for OSR baselines, we compare our method against CE+ \cite{vaze2022openset} (maximizing the closed-set accuracy of the source model) and MC-Dropout \cite{linmans2020efficient}. In addition, we adopt SOTA OSR baselines, including T3PO \cite{galdran2022test} and CSI \cite{tack2020csi} as representative of augmentation-based approaches. The former uses test-time image transformation prediction, while the latter leverages contrastive learning to contrast the sample with distributionally shifted augmentations of itself. Beyond these, our method is benchmarked against leading SF-OSDA methods, including \rev{U-SFAN \cite{roy2022uncertainty} and SHOT$^\ast$ (the open-set version of SHOT) \cite{liang2020we}.} To ensure a comprehensive assessment of $\ours$\rev{, we have also included comparisons with Co-learn$^\ast$ \cite{Zhang_2023_ICCV}, the open-set version of Co-learn model \cite{Zhang_2023_ICCV}, which, akin to ours, employs a pre-trained model for the calculation of pseudo-labels. In Co-learn$^\ast$, the SHOT$^\ast$ algorithm is integrated within Co-learn to detect open-set samples.} \rev{To ensure a fair comparison, we utilize our custom self-supervised ViT as the selected pre-trained model.} For CE+, MC-Dropout, and T3PO, we use their codebases released by \cite{galdran2022test}, suitably tailored to our experimental setting.  In contrast, for the rest, their native implementations were employed with suggested settings to secure an impartial comparison.
% we have developed \Guillaume{OS--Co-learn}—an enhanced variant of the SHOT method that leverages our custom self-supervised ViT embedding instead of relying on the source model for conducting $K$-means clustering and computing pseudo-labels. 

%Finally, for SF-OSDA baselines, we compare with SHOT \cite{liang2020we}, which is based on pseudo labels generation via $K$-means, and U-SFAN \cite{roy2022uncertainty}, which is based on uncertainty estimation. 
%Other well-known SF-TTA like TENT \cite{wang2021tent} methods strongly rely on the closed-world assumption and therefore are not applicable here. 
%Notably, these methods focus on detecting open-set categories and do not aim to tackle domain shift problems. }

\subsection{Implementation Details}
\begin{figure}[t]
    \centering
    \includegraphics[width=0.49\linewidth]{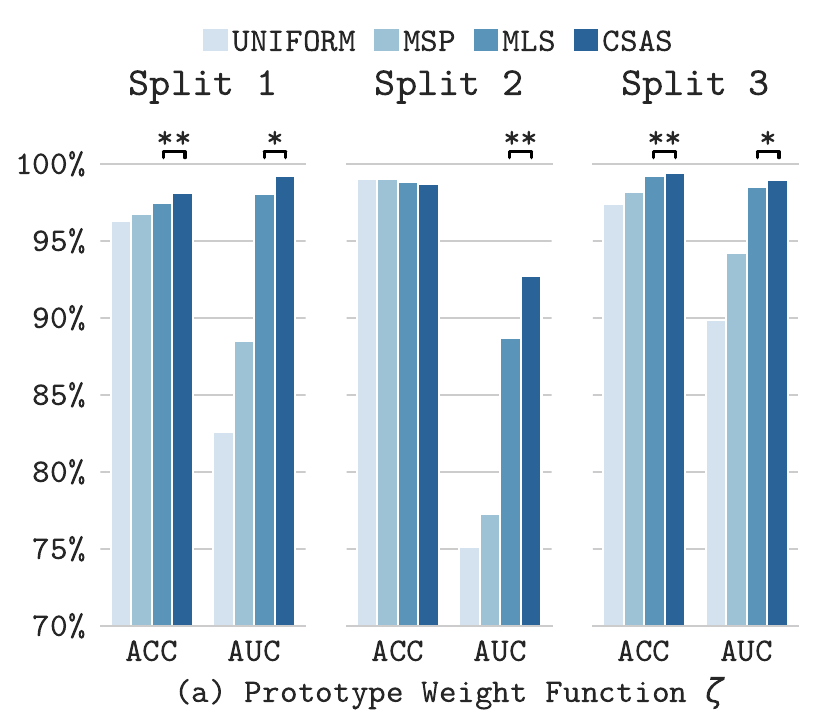}
    \includegraphics[width=0.49\linewidth]{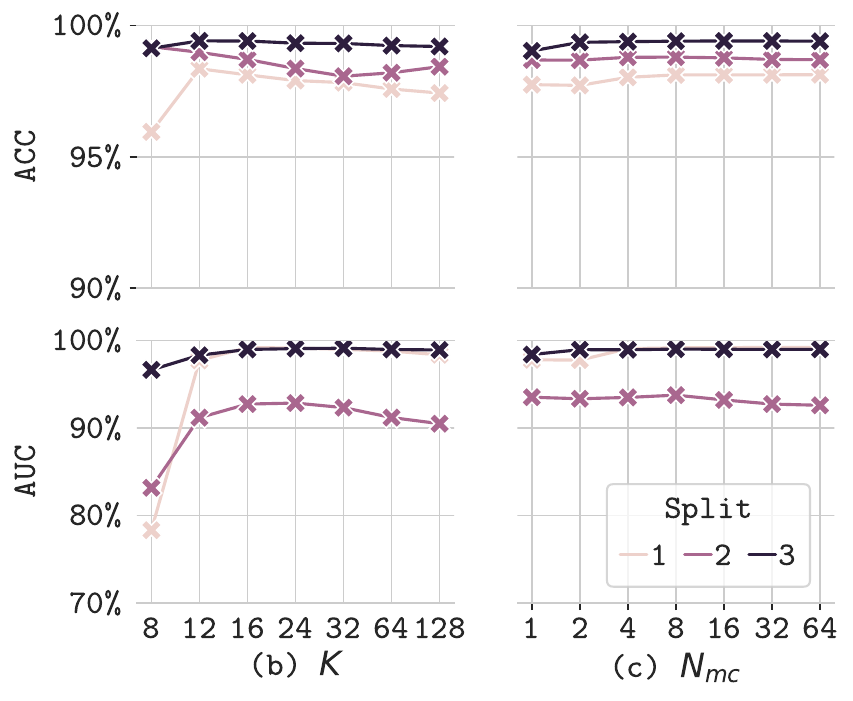}
    \caption{\textbf{Ablation studies for computing ViT guided prototypes in Kather-16$\to$Kather-19 adaptation}. \textbf{(a)} We investigate various strategies for a function used for computing class prototypes, namely: uniform weighting (UNIFORM), weighting by maximum softmax probability (MSP), positive maximum logit score weighting (MLS), and leveraging the proposed CSAS as delineated in Eq.~\ref{eq:csas_zeta}; \textbf{(b)} We scrutinize the impact of varying the cluster count $K$ within the $K$-means algorithm as utilized in CSAS; \textbf{(c)} We examine how the number of Monte-Carlo simulations $N_{mc}$
affects the robustness of CSAS. \\ \small{Paired t-test: $[**]$ $p<0.001$, $[*]$ $0.001\le p<0.1$, $[n]$ $p\ge0.1$.}}
    \label{fig:ablation_csas}
\end{figure}

We adopt the ViT/B-16 encoder as $\mathcal{F}$ starting from a self-supervised DINO \cite{caron2021emerging}  initialization and MobileNet-V2 \cite{sandler2018mobilenetv2} for the source-trained model $f_s$. We train the source model $f_s$ for 100 epochs on Kather-16, 20 epochs on Kather-19, and CRCTP, respectively, for all splits using the training strategy mentioned in CE+ \cite{vaze2022openset}. For self-training the vision transformer on the target domain $\mathcal{D}_t$, we use AdamW \cite{loshchilov2017decoupled} optimizer with a learning rate of 2.5e-4 for 40, 5, and 10 epochs on Kather-16, Kather-19, and CRCTP datasets, respectively. We compute the class prototypes using our proposed CSAS with $K=16$ clusters, $N_{\text{mc}}=32$ Monte-Carlo simulations (see ablations in Fig. \ref{fig:ablation_csas}(b)-(c)), and temperature $\tau=0.07$. Finally, for the model adaptation step, we self-train $f_s$ on the target domain with the obtained log pseudo-labels for five epochs on Kather-16 and two epochs otherwise, using Adam optimizer 
\cite{kingma2014adam}
with a learning rate of 1e-3. Our approach, $\ours$, demonstrates its versatility by being compatible with an array of open-set detection strategies. In particular, using $\ours$, we have adapted a source model $f_s$ initially trained with MC-Dropout or T3PO. While the overall adaptation framework and the hyperparameter settings remain unchanged, we make a crucial adjustment during the self-training stage on the target domain: we substitute the original cross-entropy loss with our custom $l_2$ loss function. In the following, we refer to the modified models $\ours$ (CE+), $\ours$ (MC-Dropout), and $\ours$ (T3PO), indicating the deployment of Distill-SODA on source model initially trained through CE+, MC-Dropout, and T3PO, respectively. It is important to note that the open-set scores employed for detecting open-set samples remain consistent with the respective source models. For example, CE+ utilizes the MLS score, MC-Dropout utilizes entropy measurements from 32 different predictions for uncertainty estimation, and T3PO relies on the maximum softmax probability on the augmentation prediction task. The experiments were carried out using PyTorch 1.13 on an NVIDIA GeForce GTX 1080 Ti GPU  with 12GB of memory.
%\Guillaume{Only the SSL ViT training requires two of them.}

\begin{table*}[t]
\caption{\textbf{Performance comparison against state-of-the-art methods.} We assessed performance using closed-set domain adaptation accuracy ({ACC$\uparrow$} in \%) and open-set detection  capability ({AUC$\uparrow$} in \%) for pairwise model adaptation on  Kather-16, Kather-19, and CRCTP datasets. For SF-OSDA approaches, the source model $f_s$ was trained using methods denoted in brackets. The mean row represents the overall average ACC and AUC over 18 experiments. These results have been averaged over 10 seeds. Scores where Distill-SODA outperforms other benchmarks are \underline{underlined}, and the highest scores are highlighted in \textbf{bold}. \\ $[\ddagger]\ p<0.001$, $[\dagger]\ 0.001\le p<0.1$, $[n]\ p\ge0.1$; paired t-test with respect to top baseline results.}
\label{tab:main_results}
\centering
\resizebox{\textwidth}{!}{
\begin{tabular}{@{}cc||cccc||ccc|ccc@{}}
\toprule

 \multicolumn{2}{c||}{} & 
 \multicolumn{4}{c||}{\begin{tabular}[c]{@{}c@{}}\textbf{\scriptsize Open-Set Detection Methods} \end{tabular}} & \multicolumn{6}{c}{\begin{tabular}[c]{@{}c@{}} \textbf{\scriptsize Source-Free Open-Set Domain Adaptation Methods} \end{tabular}} \\ 

\midrule                                    
\multicolumn{1}{c|}{\rotninty{\scriptsize Split}} & \rotninty{\scriptsize Metric} & 
\begin{tabular}[c]{@{}c@{}} {\scriptsize CE+} \\ \end{tabular} &
\begin{tabular}[c]{@{}c@{}} {\scriptsize MC-Dropout} \\  \end{tabular} & 
\begin{tabular}[c]{@{}c@{}} {\scriptsize T3PO} \\ \end{tabular} & 
\begin{tabular}[c]{@{}c@{}} {\scriptsize CSI} \\ \end{tabular} &
\begin{tabular}[c]{@{}c@{}} \textbf{\scriptsize Distill-SODA} \\ {\scriptsize (CE+)} \end{tabular} & 
\begin{tabular}[c]{@{}c@{}} \textbf{\scriptsize Distill-SODA} \\ {\scriptsize (MC-Dropout)} \end{tabular} & 
\begin{tabular}[c]{@{}c@{}} \textbf{\scriptsize Distill-SODA} \\ {\scriptsize (T3PO)} \end{tabular} &
\begin{tabular}[c]{@{}c@{}} {\scriptsize U-SFAN} \\ {\scriptsize (CE+)} \end{tabular} & 
\begin{tabular}[c]{@{}c@{}} {\scriptsize \rev{SHOT$^\ast$}} \\ {\scriptsize (CE+)} \end{tabular} & 
% \begin{tabular}[c]{@{}c@{}} \Guillaume{OS--Co-learn} \\ (CE+) \end{tabular} & 
\begin{tabular}[c]{@{}c@{}} {\scriptsize \rev{Co-learn$^\ast$}}\\ {\scriptsize (CE+)} \end{tabular} \\

\midrule

% \multicolumn{2}{c||}{\begin{tabular}[c]{@{}c@{}} {\scriptsize \Guillaume{Use SSL}} \\ {\scriptsize \Guillaume{Target Features}} \end{tabular}} & \xmark & \xmark & \xmark & \xmark & \cmark & \cmark & \cmark & \xmark & \xmark & \cmark \\
% \midrule

\multicolumn{12}{c}{\textbf{Kather-19} $\to$ \textbf{Kather-16} \textbf{/} \textbf{Kather-16} $\to$ \textbf{Kather-19}}\\
\midrule
\multicolumn{1}{c|}{\multirow{2}{*}{1}} & ACC
& 75.3/88.6 & 76.4/88.2 & 89.2/92.9 & 79.3/89.2
& \underline{94.3$^\ddagger$}/\underline{\textbf{97.8$^\ddagger$}} & \underline{95.0$^\ddagger$}/\underline{97.6$^\ddagger$} & \underline{\textbf{95.1$^\ddagger$}}/\underline{\textbf{97.8$^\ddagger$}}
& 89.1/87.5
& 88.5/89.6 
% & 90.8/94.6
% & 94.6/96.3 
& 89.9/90.7
\\

\multicolumn{1}{c|}{} & AUC
& 85.4/78.0 & 80.5/81.1 & 88.4/80.7 & 66.4/74.1 
& \underline{\textbf{95.7$^\ddagger$}}/\underline{\textbf{99.2$^\ddagger$}}  & \underline{92.5$^\ddagger$}/\underline{97.7$^\ddagger$} & 
85.5/\underline{92.3$^\dagger$}
& 85.3/61.1
& 80.5/51.4 
% & 82.6/53.1
% & 58.8/61.9 
& 79.0/65.9
\\

\cmidrule(l){1-12} 

\multicolumn{1}{c|}{\multirow{2}{*}{2}} & ACC
& 97.9/75.7 & 98.2/77.0 & 98.8/89.1 & 95.2/92.1
& \underline{99.3$^n$}/\underline{98.7$^\dagger$} & \underline{\textbf{99.4$^\dagger$}}/\underline{98.4$^\dagger$} & \underline{99.2$^n$/\textbf{99.1$^\ddagger$}}
& 97.7/88.1
& 97.6/90.1 
% & 98.7/97.2
% & 97.6/99.0 
& 98.4/90.6
\\

\multicolumn{1}{c|}{} & AUC
& 86.1/66.7 & 86.0/67.7 & 85.6/78.8 & 72.2/61.9
& \underline{\textbf{96.0$^\ddagger$}}/\underline{\textbf{92.1$^\ddagger$}} & \underline{91.9$^\ddagger$}/\underline{81.3$^n$} & 
\underline{87.7$^n$}/\underline{83.8$^\dagger$}
& 87.1/65.4
& 83.8/65.8 
% & 83.0/63.6
% & 49.0/71.2 
& 82.0/69.5
\\

\cmidrule(l){1-12} 

\multicolumn{1}{c|}{\multirow{2}{*}{3}} & ACC
& 84.5/73.4 & 84.9/81.4 & 88.9/91.0 & 70.1/94.0
& \underline{93.6$^\ddagger$}/\underline{\textbf{99.0$^\dagger$}} 
& \underline{93.4$^\ddagger$}/\underline{98.6$^\dagger$}
& \underline{\textbf{93.8$^\ddagger$}}/\underline{98.9$^\dagger$}
& 88.0/84.9
& 85.7/86.4 
% & 89.8/92.8
% & 95.4/96.1 
& 88.0/89.9
\\

\multicolumn{1}{c|}{} & AUC
& 83.0/65.1 & 82.2/64.5 & 85.6/78.6 & 66.1/55.8
& \underline{\textbf{95.2$^\ddagger$}}/\underline{\textbf{98.8$^\ddagger$}} & \underline{88.9$^\dagger$}/\underline{94.2$^\ddagger$} &
85.2/\underline{95.0$^\ddagger$}
& 85.0/65.7 
& 73.9/64.3 
% & 79.7/64.8
% & 48.1/76.4 
& 76.1/79.9
\\

\midrule
\multicolumn{12}{c}{\textbf{CRCTP} $\to$ \textbf{Kather-16} \textbf{/} \textbf{Kather-16} $\to$ \textbf{CRCTP}}\\
\midrule

\multicolumn{1}{c|}{\multirow{2}{*}{1}} & ACC
& 71.0/67.3 & 72.1/68.4 & 72.0/76.5 & 66.4/61.9
& \underline{95.2$^\ddagger$}/\underline{\textbf{80.2$^\ddagger$}} & \underline{94.8$^\ddagger$}/\underline{80.0$^\ddagger$} & \underline{\textbf{95.6$^\ddagger$}}/\underline{79.9$^\ddagger$} 
& 84.4/74.1
& 87.8/76.0 
% & 88.8/78.3
% & 92.4/80.4 
& 89.9/76.6
\\

\multicolumn{1}{c|}{} & AUC
& 69.6/67.9 & 64.9/70.6 & 82.8/60.8 & 80.9/65.0
& \underline{\textbf{94.4$^\ddagger$}}/\underline{\textbf{80.4$^\ddagger$}} & \underline{85.5$^\dagger$}/\underline{72.3$^\dagger$} & 
77.5/\underline{73.1$^n$}  
& 76.2/61.6
& 78.7/61.7 
% & 80.1/58.3
% & 49.6/64.7 
& 80.1/62.0
\\

\cmidrule(l){1-12} 

\multicolumn{1}{c|}{\multirow{2}{*}{2}} & ACC
& 97.1/85.2 & 96.8/84.2 & 81.4/92.8 & 95.2/84.3
& \underline{99.2$^\dagger$}/\underline{\textbf{95.8$^\dagger$}} & \underline{\textbf{99.8$^\ddagger$}}/\underline{95.7$^\dagger$} & \underline{99.1$^n$}/\underline{95.6$^n$}
& 98.0/94.4
& 97.5/95.0 
% & 97.9/95.3
% & 96.0/96.0 
& 97.7/95.3
\\

\multicolumn{1}{c|}{} & AUC
& 76.9/70.2 & 75.0/69.0 & 72.6/70.7 & 72.2/61.4
& \underline{\textbf{92.3$^\ddagger$}}/\underline{\textbf{89.8$^\ddagger$}} & \underline{84.3$^\ddagger$}/\underline{86.6$^\ddagger$} & \underline{77.1$^n$}/\underline{84.6$^\ddagger$} 
& 75.2/77.9
& 72.8/76.0 
% & 74.6/70.7
% & 50.2/82.8 
& 73.5/79.6
\\

\cmidrule(l){1-12} 

\multicolumn{1}{c|}{\multirow{2}{*}{3}} & ACC
& 89.0/66.9 & 86.3/70.0 & 83.7/81.9 & 70.1/59.2
& \underline{95.9$^n$}/\underline{84.3$^n$} & \underline{96.0$^\dagger$}/\underline{\textbf{84.4$^n$}} & \underline{\textbf{96.5$^\dagger$}}/\underline{84.2$^n$} 
& 95.0/82.0
& 95.3/82.7 
% & 95.1/84.2
% & 90.8/84.8 
& 95.4/82.8
\\

\multicolumn{1}{c|}{} & AUC
& 73.1/69.4 & 66.4/68.4 & 82.1/70.3 & 66.1/64.3
& \underline{\textbf{89.2$^\dagger$}}/\underline{\textbf{87.6$^\ddagger$}} & 76.6/\underline{81.6$^\dagger$}  & 
\underline{84.7$^n$}/\underline{74.6$^n$}  
& 72.5/72.1
& 71.9/71.8 
% & 72.5/69.6
% & 39.3/75.0 
& 71.7/73.5
\\

\midrule
\multicolumn{12}{c}{\textbf{CRCTP} $\to$ \textbf{Kather-19} \textbf{/} \textbf{Kather-19} $\to$ \textbf{CRCTP}}\\
\midrule
\multicolumn{1}{c|}{\multirow{2}{*}{1}} & ACC 

& 73.9/52.5 & 75.4/52.5 & 75.6/53.1 & 72.2/60.7
& \underline{\textbf{96.8$^\ddagger$}}/\underline{\textbf{80.5$^\ddagger$}} & \underline{96.7$^\ddagger$}/\underline{80.3$^\ddagger$} & \underline{96.3$^\ddagger$}/\underline{80.1$^\ddagger$} 
& 89.1/73.7
& 84.6/74.7 
% & 92.4/78.6
% & 93.9/80.3
& 86.2/75.8
\\

\multicolumn{1}{c|}{} & AUC 

& 72.0/63.9 & 76.7/63.6 & 85.9/61.8 & 72.5/64.4
& \underline{\textbf{98.3$^\ddagger$}}/\underline{\textbf{78.4$^\ddagger$}} & \underline{93.7$^\ddagger$}/\underline{70.4$^\dagger$} & \underline{95.8$^\ddagger$}/\underline{67.7$^n$}
& 56.9/52.6
& 66.4/56.7 
% & 58.3/48.5
% & 65.5/58.6 
& 74.1/59.0
\\

\cmidrule(l){1-12} 

\multicolumn{1}{c|}{\multirow{2}{*}{2}} & ACC 

& 95.2/88.6 & 95.8/88.5 & 94.9/87.5 & 93.9/93.1
& \underline{\textbf{99.4$^\ddagger$}}/\underline{\textbf{95.8$^\ddagger$}} & \underline{99.1$^\ddagger$}/\underline{95.6$^\dagger$} & \underline{99.0$^\ddagger$}/\underline{95.7$^\dagger$}
& 90.6/94.4
& 92.2/94.1 
% & 97.3/95.1
% & 98.9/95.8 
& 93.5/94.5
\\

\multicolumn{1}{c|}{} & AUC 

& 73.6/70.6 & 73.4/72.0 & 74.0/68.2 & 71.3/77.1 
& \underline{\textbf{95.1$^\ddagger$}}/\underline{\textbf{90.2$^\ddagger$}} & \underline{91.3$^\ddagger$}/\underline{86.4$^\ddagger$} & \underline{79.2$^\ddagger$}/\underline{83.6$^\ddagger$}
& 58.9/71.4
& 68.6/78.3 
% & 53.4/72.5
% & 65.4/80.5 
& 76.0/82.8
\\

\cmidrule(l){1-12} 

\multicolumn{1}{c|}{\multirow{2}{*}{3}} & ACC 

& 94.4/57.7 & 94.3/57.2 & 94.9/59.7 & 94.9/62.1
& \underline{\textbf{99.1$^\ddagger$}}/\underline{85.1$^\dagger$} & \underline{98.9$^\ddagger$}/\underline{\textbf{85.3$^\dagger$}} & \underline{98.6$^\ddagger$}/\underline{84.2$^\dagger$}
& 92.2/82.9
& 93.9/80.9 
% & 95.3/82.1
% & 98.9/84.6 
& 95.9/81.4
\\

\multicolumn{1}{c|}{} & AUC 

& 73.5/79.2 & 75.9/74.3 & 81.9/77.9 & 85.3/78.4
& \underline{\textbf{97.9$^\ddagger$}}/\underline{\textbf{84.3$^\ddagger$}} & \underline{97.1$^\ddagger$}/78.0 & \underline{92.9$^\ddagger$}/\underline{81.4$^n$} 
& 61.1/62.8
& 68.7/69.5 
% & 55.8/62.6
% & 73.6/69.4 
& 78.0/73.0
\\

\midrule
\midrule

\multicolumn{1}{c|}{\multirow{2}{*}{Mean}} & ACC 
& 79.7 & 80.4 & 83.6 & 79.7 & \underline{\textbf{93.9}}\posmargin{14.2} & \underline{93.8}\posmargin{13.4} & \underline{93.8}\posmargin{10.2} & 88.1\posmargin{8.4} & 88.5\posmargin{8.8} 
% & 91.4\posmargin{11.7} 
% &  92.9\posmargin{13.2}
& 89.6\posmargin{9.9}
\\
 
\multicolumn{1}{c|}{} & AUC 
& 73.6 & 72.9 & 77.0 & 69.7 & \underline{\textbf{91.9}}\posmargin{18.3} & \underline{86.1}\posmargin{13.2} & \underline{83.4}\posmargin{6.4} & 69.4\negmargin{4.2} & 70.0\negmargin{3.6} 
% & 66.9\negmargin{6.7} 
% & 63.3\negmargin{10.3}
& 74.2\posmargin{0.6}
\\

\bottomrule
\end{tabular}}
\end{table*}

\begin{figure*}[t]
    \centering
    \includegraphics[width=1.0\linewidth]
    {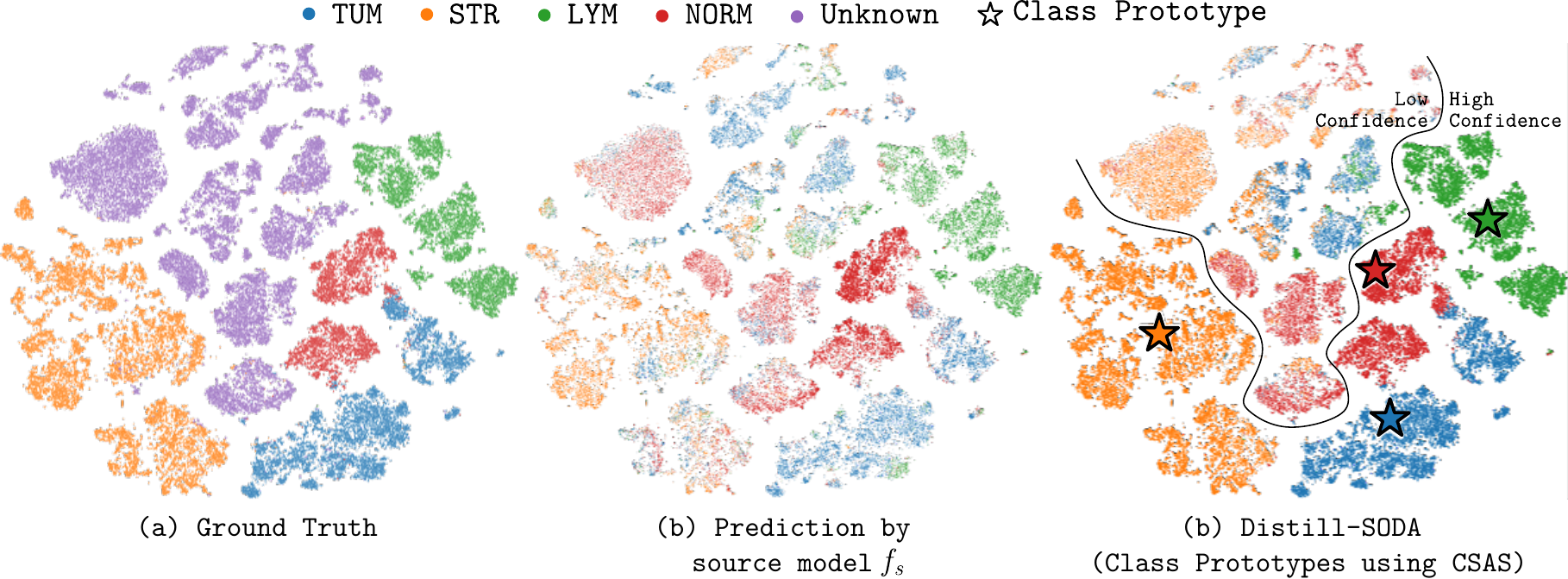}
    \caption{\textbf{The t-SNE visualization comparison of feature embeddings obtained from self-supervised trained transformer encoder $\mathcal{F}$:} This figure showcases the distinctions among tissue types within the target domain images (Kather-19, Split 1) using t-SNE \cite{van2008visualizing}. Each tissue type is uniquely color-coded for clarity: 
    \textbf{(a)} Reflects the actual labels; \textbf{(b)} Illustrates the feature embedding from the BatchNorm re-calibrated source model $f_s$ trained on Kather-16; and \textbf{(c)} Depicts the feature embedding from our proposed method, $\ours$. The weighted average class prototypes are obtained in the transformer’s embedding space, with the intensity of the color signifying the confidence level of the labels.} 
    %    \textit{Labels' confidence} $\propto$ \textit{Color Density
    %\textbf{(c)} SSL ViT-guided class prototypes obtained from classwise CRMLS weighted average features.(Eq. \ref{eq:class_proto})%\Guillaume{The pseudo-labels in (b) are noisy concerning the ground truth and overall unconfident for both open-set and closed-set samples, making open-set examples hardly detectable. However, the pseudo-labels of our method in (c) match the ground truth. The pseudo-labels are more confident for closed-set examples than open-set examples, allowing accurate detection of Open-Set categories based on confidence.}
    \label{fig:tsne}
\end{figure*}

% \subsection{Results and Ablation Studies}
\subsection{Comparisons with State-of-the-Art Methods} 
Table \ref{tab:main_results} compares $\ours$ against SOTA methods for closed-set domain adaptation scenario via ACC metric and open-set recognition capability through AUC metric on three splits for adapting source model initially trained through CE+, MC-Dropout and T3PO across various target domains. Remarkably, $\ours$ (CE+) consistently outperforms all the baselines, exhibiting the highest AUC and ACC scores for all three CRC tissue datasets in all adaptation scenarios. Notably, $\ours$ (CE+) demonstrates a significant performance gain in open-set recognition, achieving a remarkable +18.3\% increase in AUC and exhibits a notable substantial +14.2\% increase in ACC for closed-set domain adaptation, averaged over 18 experimental runs when compared to the original CE+ method. Several conclusions can be drawn based on the experimental comparison to other methods. First, one can observe that open-set detection methods lag behind $\ours$ when exposed to covariate shifts due to variations in tissue's visual appearances. Furthermore, $\ours$ (MC-Dropout) and $\ours$ (T3PO) consistently demonstrate significant average improvements over their respective OSR models.  When compared to $\ours$ (CE+), their final mean accuracy is only marginally lower by $-0.1\%$, but their Mean AUC scores exhibit more pronounced differences, lagging behind $\ours$ (CE+) by $5.8\%$ and $8.5\%$, respectively. This can be attributed to their reliance on prediction entropy and augmentation detection as open-set scores instead of MLS utilized by CE+. In summary, it is evident that while $\ours$ can effectively adapt OSR methods to the target domain, yet the best results arise from pairing a simple CE+ model with $\ours$. This is a favorable outcome as it simplifies the training and model adaptation process, particularly in medical applications.
Second, similar poor performance trends are observed for recent SF-OSDA methods, including U-SFAN and \rev{SHOT$^\ast$}. Both \rev{SHOT$^\ast$} and U-SFAN detect open-set samples by analyzing the source model's features, which may not be very informative for open-set detection to handle severe chromatic shifts of tissue patches. We argue that the superior performance of $\ours$ is a direct effect induced by knowledge distillation through a separate self-supervised trained vision transformer in the distributionally shifted unlabeled target domain.  Nonetheless, it is noteworthy that $\ours$ (CE+) consistently surpasses the \rev{Co-learn$^\ast$} baseline, despite the latter also benefiting from our vision transformer (with an average increase of \rev{4.3\%} in ACC and \rev{17.7\%} in AUC). Therefore, we conclude our approach for generating class prototypes within the target domain, followed by the adaptation of the source model for open-set scenarios, proves to be more effective than the \rev{Co-learn$^\ast$} method\rev{, which merges SHOT$^\ast$ with Co-learn strategy.}

The quantitative results are supported by the t-SNE \cite{van2008visualizing} visualization results (Fig. \ref{fig:tsne}) of self-supervised trained ViT encoder $\mathcal{F}$ of feature embeddings on target domain images (Kather-19, Split 1). Our transformer-guided class prototypes successfully refine the weak pseudo-labels (given by $f_s$) of the target domain images in the ViT encoder's feature space with reliable confidence in the presence of open-set samples and covariate shift caused by changes in tissue color appearance.
%
%

% \vspace{0.5em}
% \noindent
% \textbf{$\ours$ has been shown to integrate effectively with various pre-trained self-supervised ViTs.} In Table~\ref{tab:ablation_vision_models}, we present the performance of $\ours$ when using the embedding space of a pre-trained self-supervised vision transformer, such as CTransPath~\cite{wang2022}, or one self-trained on a public histopathology dataset or the target dataset using our proposed adversarial style augmentation. The third dataset, distinct from the source and target domains, serves as the public dataset in our experiments. Notably, in the first two cases, the vision transformer has no prior exposure to target domain images, leading to a slightly lower average performance compared to the vision transformer self-trained on the target dataset. Additionally, a compelling observation is that the vision transformer self-trained using DINO with our proposed adversarial style augmentation on the public dataset competes with the pre-trained CTransPath model's performance.

\begin{figure}[t]
    \centering
    \includegraphics[width=\linewidth]{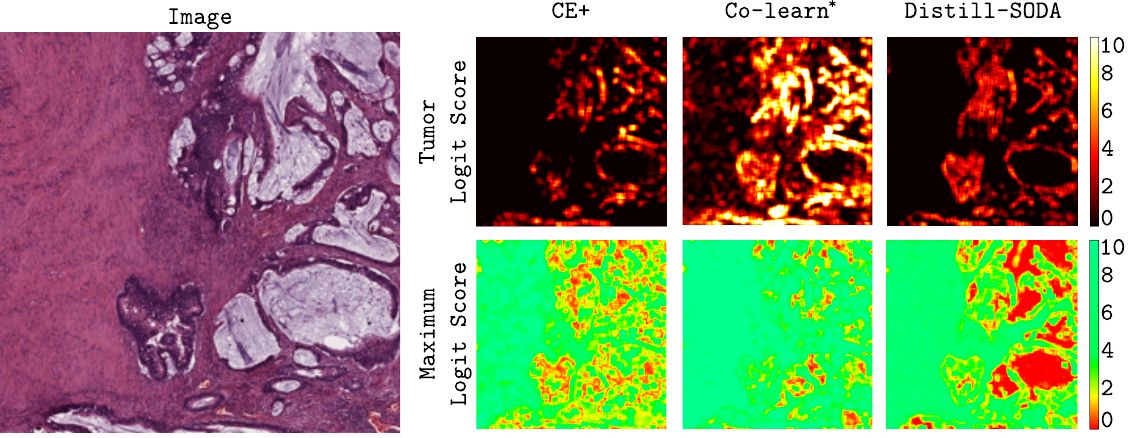}
    \caption{\textbf{Tissue image segmentation results} for the Kather-19$\to$Kather-16 \rev{(Split 1)} model adaptation performed by CE+, \rev{Co-learn$^\ast$}, and $\ours$ on unseen tissue image: Visualizations showcase the (top) predicted logit score for the tumor (TUM) class; (bottom) the predicted MLS score with green indicating regions identified as known tissue classes (TUM, STR, LYM, \rev{NORM}), and red denoting areas classified into novel, open-set categories (c-STR, DEB, ADI and BACK).}
    \label{fig:segmentation}
\end{figure}

\begin{table}[t]
    \centering
    \caption{\textbf{Ablation on types of distillation loss} for adapting source model $f_s$ to target domain (Kather-16 $\to$ Kather-19) using pseudo-logits of Eq. \ref{eq:self_distill}. We compare the performance in terms of ACC and AUC metrics of the target training pseudo-labels with that of the adapted model $f_t$ trained with cross-entropy (CE), SHOT, $l_1$, and $l_2$ on the target test set. {\color{red}Red} denotes severe degradation while {\color{cyan}cyan} denotes minor change.
    % The suggested $l_2$ loss is the only objective ensuring $f_t$ does not suffer a performance drop on unseen test target data.
    }
    \resizebox{\linewidth}{!}{
    \begin{tabular}{c|c|c|llll}
    \toprule

    \multirow{2}{*}{\rotninty{Split}} & \multirow{2}{*}{\rotninty{Metric}} & \multirow{2}{*}{\begin{tabular}[c]{@{}c@{}}target\\ \
{\scriptsize pseudo-}\\{\scriptsize labels} \end{tabular}}&\multicolumn{4}{|c}{Distillation loss} \\

    \cmidrule{4-7}

    & & & CE & SHOT & $l_1$ & $l_2$ \\

    \midrule

    \multirow{2}{*}{1} & ACC & 98.1 & 97.3\color{cyan}(-0.8) & 96.8\color{red}(-1.3) & 97.4\color{cyan}(-0.7) & 98.0\color{cyan}(-0.1)\\
    & AUC & 99.2 & 80.1\color{red}(-19.1) & 81.2\color{red}(-18) & 98.8\color{cyan}(-0.4) & 99.3\color{cyan}(+0.1) \\
    \midrule

    \multirow{2}{*}{2} & ACC & 98.7 & 98.7\color{cyan}(-0.0) & 98.5\color{cyan}(-0.2) & 98.6\color{cyan}(-0.1) & 98.7\color{cyan}(-0.0) \\
    & AUC & 92.7 & 68.2\color{red}(-25.5) & 68.6\color{red}(-24.1) & 87.8\color{red}(-4.9) & 92.2\color{cyan}(-0.5) \\
    \midrule

    \multirow{2}{*}{3} & ACC & 99.4 & 98.4\color{red}(-1.0) & 98.1\color{red}(-1.3) & 98.4\color{red}(-1.0) & 99.0\color{cyan}(-0.4) \\
    & AUC & 99.0 & 86.2\color{red}(-12.8) & 85.3\color{red}(-13.7) & 98.4\color{cyan}(-0.6) & 98.8\color{cyan}(-0.2) \\
    
    \bottomrule
    \end{tabular}}
    \label{tab:loss}
\end{table}

\begin{table*}[t]
\centering
\caption{\rev{\textbf{Comparison of pre-trained and self-trained vision transformers for $\ours$}: This comparison charts the efficacy of utilizing a pre-trained self-supervised model like CTransPath~\cite{wang2022}, MAE \cite{he2022masked}, DINO \cite{caron2021emerging}, and MoCoV3 \cite{fan2021multiscale}, against a self-trained vision transformer using DINO with our proposed adversarial style augmentations. The evaluation is conducted on both a \textbf{target} ($\mathcal{D}_t$) and a separate but similar \textbf{public} ($\mathcal{D}_p$) dataset (excluding target data). Results are averaged over the three separate splits.}}
\label{tab:ablation_vision_models}
\resizebox{\textwidth}{!}{
\begin{tabular}{@{}c|c|c|c|c|cc|cc|cc|cc|cc|cc||cc@{}}
\toprule
\multirow{3}{*}{\rotninty{Method}} & \multirow{3}{*}{\begin{tabular}[c]{@{}c@{}} SSL \\ Method \end{tabular}} & \multirow{3}{*}{\begin{tabular}[c]{@{}c@{}} SSL \\ Backbone \end{tabular}} & \multirow{3}{*}{\begin{tabular}[c]{@{}c@{}} SSL \\ Pre-Training \\ Dataset \end{tabular}} & 
\multirow{3}{*}{\begin{tabular}[c]{@{}c@{}} SSL \\ Self-Training \\ Dataset \end{tabular}} &
  
  \multicolumn{2}{c|}{\begin{tabular}[c]{@{}c@{}} \textbf{\scriptsize Kather-19} \\ \textbf{\scriptsize $\downarrow$} \\ \textbf{\scriptsize Kather-16}  \end{tabular}} &
  \multicolumn{2}{c|}{\begin{tabular}[c]{@{}c@{}} \textbf{\scriptsize CRCTP} \\ \textbf{\scriptsize $\downarrow$} \\ \textbf{\scriptsize Kather-16}  \end{tabular}} &
  \multicolumn{2}{c|}{\begin{tabular}[c]{@{}c@{}} \textbf{\scriptsize Kather-16} \\ \textbf{\scriptsize $\downarrow$} \\ \textbf{\scriptsize Kather-19}  \end{tabular}} &
  \multicolumn{2}{c|}{\begin{tabular}[c]{@{}c@{}} \textbf{\scriptsize CRCTP} \\ \textbf{\scriptsize $\downarrow$} \\ \textbf{\scriptsize Kather-19}  \end{tabular}} &
  \multicolumn{2}{c|}{\begin{tabular}[c]{@{}c@{}} \textbf{\scriptsize Kather-16} \\ \textbf{\scriptsize $\downarrow$} \\ \textbf{\scriptsize CRCTP}  \end{tabular}} &
  \multicolumn{2}{c||}{\begin{tabular}[c]{@{}c@{}} \textbf{\scriptsize Kather-19} \\ \textbf{\scriptsize $\downarrow$} \\ \textbf{\scriptsize CRCTP}  \end{tabular}} & 
  \multicolumn{2}{c}{\scriptsize \textbf{Mean}} \\ 
  \cmidrule(l){6-19} 
  
   & & & & & {\scriptsize ACC}  & {\scriptsize AUC}  & {\scriptsize ACC}  & {\scriptsize AUC}  & {\scriptsize ACC}  & {\scriptsize AUC}  & {\scriptsize ACC} & {\scriptsize AUC} & {\scriptsize ACC}  & {\scriptsize AUC}  & {\scriptsize ACC}  & {\scriptsize AUC}  & {\scriptsize ACC}  & {\scriptsize AUC} \\ 

   \midrule 

   \rotninty{CE+} & \multicolumn{4}{c|}{-} & 85.9 & 84.8 & 85.7 & 73.2 & 79.2 & 69.9 & 87.8 & 73.0 & 73.1 & 69.2 & 66.3 & 71.2 & 79.7 & 73.6 \\
   
    \midrule

    \multirow{6}{*}{\rotninty{\begin{tabular}[c]{@{}c@{}} \textbf{Distill-SODA} \\ (CE+) \end{tabular}}} & {\scriptsize MAE} & {\scriptsize{ViT-B/16}} & {\scriptsize{ImageNet-1k}} & - & 76.5 & 76.1 & 65.6 & 73.6 & 59.9 & 79.0 & 59.9 & 77.3 & 73.6 & 66.9 & 72.6 & 68.1 & 68.0 \negmargin{11.7} & 73.5 \negmargin{0.1} \\
    
    & {\scriptsize MoCoV3} & {\scriptsize{ViT-B/16}} & {\scriptsize{ImageNet-1k}} & - & 94.5 & 91.8 & 92.6 & 85.6 & 94.0 & 87.9 & 94.5 & 92.4 & 85.1 & 65.7 & 85.6 & 65.0 & 91.1 \posmargin{11.4} & 81.4 \posmargin{7.8} \\
    
    & {\scriptsize DINO} & {\scriptsize{ViT-B/16}} & {\scriptsize{ImageNet-1k}} &- & 94.7 & 91.5 & 92.8 & 85.7 & 95.0 & 88.8 & 94.8 & 92.3 & 85.9 & 68.0 & 85.8 & 65.4 & 91.5 \posmargin{11.8} & 82.0 \posmargin{8.4} \\
    
    & {\scriptsize CTransPath} & {\scriptsize CTransPath} & {\scriptsize{CTransPath}} &- & 95.4 & 95.2 & 96.4 & 91.4 & 92.3 & 92.3 & 96.1 & 96.9 & 86.6 & 79.8 & \textbf{87.2} & 80.5 & 92.4 \posmargin{12.7} & 89.4 \posmargin{15.8} \\

    \cmidrule{2-19}
    
    % {\scriptsize DINO} & public \\
    & {\scriptsize DINO+AdvStyle} & {\scriptsize{ViT-B/16}} & {\scriptsize{ImageNet-1k}} &  {\scriptsize{$\mathcal{D}_p$}}  & \textbf{96.3} & 93.0 & \textbf{96.8} & \textbf{93.3} & 95.9 & 92.1 & 97.2 & 94.1 & \textbf{87.5} & 84.4 & 83.8 & 69.2 & 93.0 \posmargin{13.3} & 87.7 \posmargin{14.1} \\

    % \cmidrule{2-19}
    
    % {\scriptsize DINO} & target & & & & & 98.0 & 94.2 & 97.6 & 95.6 \\
    & \textbf{{\scriptsize DINO+AdvStyle}} & {\scriptsize{ViT-B/16}} & {\scriptsize{ImageNet-1k}} &  {\scriptsize{$\mathcal{D}_t$}} & 95.7 & \textbf{95.6} & \textbf{96.8} & 92.0 & \textbf{98.5} & \textbf{96.7} & \textbf{98.4} & \textbf{97.1} & 86.8 & \textbf{85.9} & 87.1 & \textbf{84.3} & \textbf{93.9} \posmargin{14.2} & \textbf{91.9} \posmargin{18.3} \\

\bottomrule
\end{tabular}%
}
\end{table*}

\vspace{0.5em}
\noindent
\textbf{Tissue image segmentation.} To demonstrate the capability of our approach for segmentation on large-resolution tissue-level images, we adapted the patch-level classifier for the Kather-19$\to$Kather-16 model adaptation. In Fig. \ref{fig:segmentation}, we show our method's pixel-level segmentation results on a large tissue image (5000$\times$5000 pixels) alongside segmentation maps generated by CE+ and \rev{Co-learn$^\ast$}. For each method, we present the segmentation maps computed using the predicted logit score for tumor (TUM) \rev{detection} and MLS \rev{for open-set recognition}. In the absence of ground truth information for the presented image, our method is seen to better segregate closed-set tissue classes (TUM, STR, LYM, NORM) from open-set classes (c-STR, DEB, ADI, BACK) (Fig. \ref{fig:segmentation}, bottom). In addition, $\ours$ produces a less noisy segmentation map, yielding better delineation of tumor regions than baselines (Fig. \ref{fig:segmentation}, top).
% To demonstrate the capability of our approach for segmentation on large-resolution tissue-level images, we adapted the patch-level classifier for the Kather-19$\to$Kather-16 model adaptation. In Fig. \ref{fig:segmentation}, we show our method's pixel-level segmentation results on a large tissue image (5000$\times$5000 pixels) alongside segmentation maps generated by competing methods, CE+ and \rev{Co-learn$^\ast$}. For each method, we present the segmentation maps computed using the predicted logit score for tumor (TUM) and MLS score for open-set and closed-set tissue classes. In the absence of ground truth information for the presented image, our method is seen to better segregate closed-set tissue classes (TUM, STR, LYM, MUC) from open-set classes (c-STR, DEB, ADI, and BACK) (Fig. \ref{fig:segmentation}, bottom). In addition, $\ours$ produces a less noisy segmentation map, yielding better delineation of tumor regions than other methods (Fig. \ref{fig:segmentation}, top).

%

%

\begin{table}[t]
\caption{\rev{\textbf{Comparative analysis of our approach against the state-of-the-art SFDA methods.} We assess the performance of our approach against CE+ \cite{vaze2022openset}, BN \cite{ioffe2015batch}, SHOT \cite{liang2020we}, and Co-learn \cite{Zhang_2023_ICCV} in the closed-set setting on the Kather-16$\to$Kather-19 and Kather-19$\to$Kather-16 adaptation tasks. In this setting, the source model $f_s$ has been trained to classify seven tissue types: (1) TUM, (2) (MUS+STR), (3) LYM, (4) NORM, (5) (DEB+MUC), (6) ADI, and (7) BACK. The results, shown as accuracy percentages on the test set,  averaged over 10 seeds.}}
\centering
\label{tab:closed}
\resizebox{\linewidth}{!}{
\begin{tabular}{c||c|ccc|c}
\toprule
     {\scriptsize Dataset} 
     & {\scriptsize CE+} 
     
     & \begin{tabular}[c]{@{}c@{}} {\scriptsize BN} \\ 
     {\scriptsize (CE+)}  \end{tabular}
     & \begin{tabular}[c]{@{}c@{}} {\scriptsize SHOT} \\ {\scriptsize (CE+)}  \end{tabular} 
     & \begin{tabular}[c]{@{}c@{}} {\scriptsize Co-learn} \\ {\scriptsize (CE+)}  \end{tabular} 
     & \begin{tabular}[c]{@{}c@{}} \textbf{\scriptsize Distill-SFDA} \\ {\scriptsize (CE+)}  \end{tabular} 
     \\
 \midrule 
     {\begin{tabular}[c]{@{}c@{}} \textbf{\scriptsize Kather-16} \\ \textbf{\scriptsize $\to$} \textbf{\scriptsize Kather-19}  \end{tabular}} & 67.6 & 85.6& 89.1 & \textbf{96.2} & 95.7 \\
 \midrule
     {\begin{tabular}[c]{@{}c@{}} \textbf{\scriptsize Kather-19} \\ \textbf{\scriptsize $\to$} \textbf{\scriptsize Kather-16}  \end{tabular}} & 62.1 & 86.6 & 87.4 & \textbf{91.4} & 90.4 \\
 \midrule
 \midrule
    Mean & 64.9 & 86.1 & 88.3 & \textbf{93.8} & 93.1 \\
\bottomrule
\end{tabular}
}
\end{table}

\subsection{Ablation Study}
\noindent
\textbf{{Weighted average class prototypes, sensitivity to $K$-means initialization/clusters.}} In Fig. \ref{fig:ablation_csas}(a), we show the effectiveness of our CSAS-based weighting function of Eq.~\ref{eq:csas_zeta} for computing weighted average class prototypes of Eq. \ref{eq:class_prototypes_gen} in the contextualized embedding space of ViT by comparing it against uniform weighting (UNIFORM), MSP, and MLS \cite{vaze2022openset}. We observe that CSAS produces superior-quality pseudo-labels. Moreover, since $\ours$ is based on $K$-means clustering, which is highly sensitive to initialization and $K$, we perform a sensitivity test on the number of Monte-Carlo runs $N_{mc}$ of $K$-\rev{m}eans and the number of clusters $K$ in Fig. \ref{fig:ablation_csas}(b)-(c). We observe that $\ours$ is insensitive to these parameters after a certain threshold, e.g., $K=12$, $N_{mc}=4$. 

\vspace{0.5em}
\noindent
\textbf{Distillation loss for source model adaptation.}
In this ablation study, we investigate different distillation loss functions for adapting the source model $f_s$ on the training set of the target domain using the corresponding prototypical pseudo-logits of Eq. \ref{eq:self_distill}. In Table \ref{tab:loss}, we report the performance of the model $f_t*$ on the test set of the target domain adapted using traditional cross-entropy loss (CE), SHOT\cite{liang2020we}'s InfoMax with CE, $l_1$ and the suggested $l_2$ loss for self-distillation. All distillation losses give similar ACC; however, only $l_1$ and $l_2$ provide high AUC scores. This is attributed to the fact that $l_1$ and $l_2$ losses also penalize the magnitude of the logits given by the target-adapted model, thus better preserving the MLS-based confidence ranking of the target domain images.

\vspace{0.5em}
\noindent
\rev{\textbf{Our method's efficacy to closed-set settings.}}
\rev{Despite being designed for the SF-OSDA setting, our method also demonstrates effectiveness in the closed-set setting. To illustrate this, we introduce a new split (fourth split) in our experiments for Kather-16 and Kather-19 datasets, focusing on seven tissue closed-set classes: (1) TUM, (2) MUS+STR, (3) LYM, (4) NORM, (5) DEB+MUC, (6) ADI, and (7) BACK. Note that the class c-STR in Kather-16 is excluded from this experiment. We re-conducted the Kather-16$\to$Kather-19 and Kather-19$\to$Kather-16 adaptation experiments accordingly. As depicted in Table \ref{tab:closed}, the closed-set version of our method, named Distill-SFDA, significantly improves upon the source model performance (CE+) with a noteworthy mean improvement of 28.2\% in accuracy. Furthermore, it outperforms well-known SFDA methods such as BN \cite{ioffe2015batch} and SHOT \cite{liang2020we} by a substantial margin of 7\% and 4.9\%, respectively. Finally, Distill-SFDA competes closely with the latest SFDA leader, Co-learn \cite{Zhang_2023_ICCV}, which also utilizes \rev{our custom self-supervised ViT} to refine pseudo-labels, showing only a marginal loss of 0.7\% in performance. The slight edge in Co-learn's performance can be attributed to re-computing pseudo-labels after every epoch, progressively refining their quality. In contrast, our method computes pseudo-labels only once, favoring a quicker training cycle.}
% self-supervised target features
%
\begin{figure}[th]
    \centering
    \includegraphics[width=0.49\linewidth]{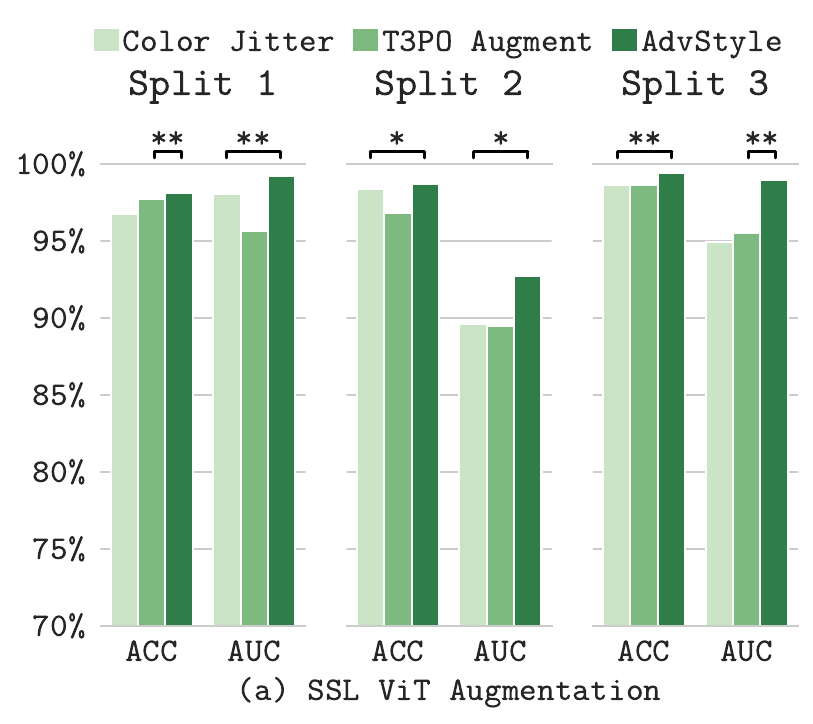}
    \includegraphics[width=0.49\linewidth]{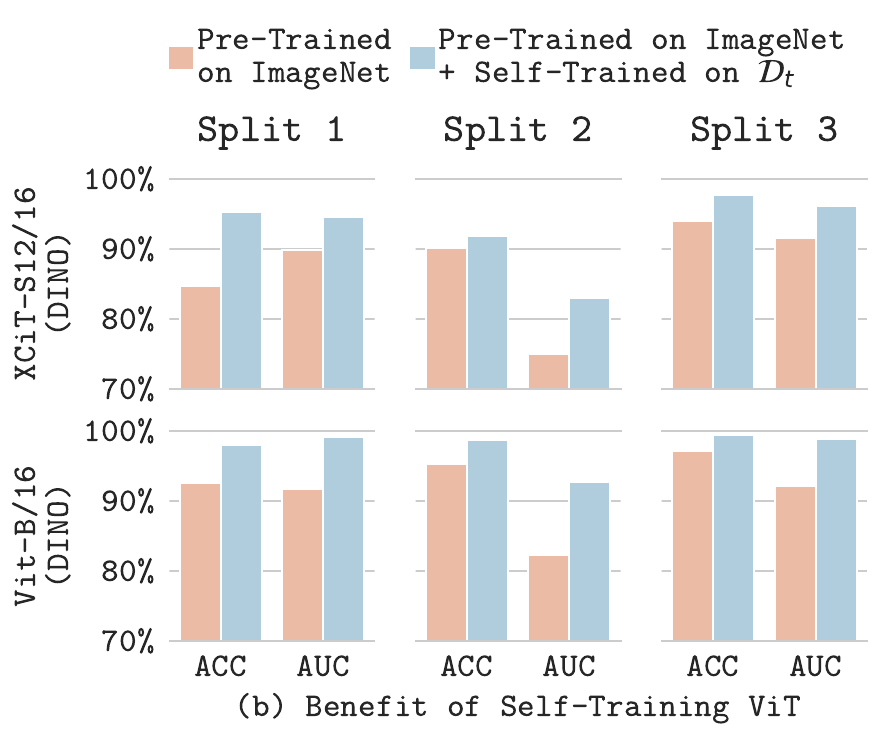}
    \caption{\textbf{Ablation studies for ViT self-training on target domain for the Kather-16$\to$Kather-19 adaptation}. \textbf{(a)} Examining style-based augmentation types used for target domain self-training of ViT on the target domain: Random Color-Jittering, T3PO's \cite{galdran2022test}, and our AdvStyle; \textbf{(b)} Comparison of pseudo-labels accuracy (ACC) and area under the curve (AUC) with and without our proposed self-training for various ViT architectures, including ViT-B/16, and XCiT-S12/16 pre-trained on ImageNet with DINO. \\ \small{Paired t-test: $[**]$ $p<0.001$, $[*]$ $0.001\le p<0.1$, $[n]$ $p\ge0.1$.}}
    \label{fig:ablation_self_training}
\end{figure}

\vspace{0.5em}
\noindent
\rev{\textbf{Evaluating the impact of knowledge distillation from pre-trained and self-trained vision transformers in  Distill-SODA framework.}}
\rev{We have also conducted a comprehensive ablation study to ascertain whether the superiority of our framework is attributable to the knowledge distillation process involving a self-supervised ViT, or if it is predominantly driven by the use of the DINO approach. To this end,  we evaluated various self-supervised ViTs pre-trained on ImageNet-1k, including MAE, MoCoV3, and the CTransPath model, which is pre-trained \rev{on the CTransPath built-in dataset \cite{wang2022continual}, a large-scale histopathological dataset comprising 15M images.} As shown in Table \ref{tab:ablation_vision_models}, the results are revealing. Except for the MAE model, knowledge distillation from all tested ViTs significantly boosts both closed-set accuracy and open-set detection compared to the baseline source model (CE+). The most marked improvements are with DINO and CTransPath, the latter benefiting from domain-specific pre-training. Additionally, applying our adversarial style augmentations in self-training vision transformers on either the target domain or a separate but similar public domain further enhances the performance. This confirms the robustness of our Distill-SODA framework, leveraging both domain-specific training and advanced self-supervised learning techniques.}

\begin{figure}[t]
    \centering
    \includegraphics[width=\linewidth]{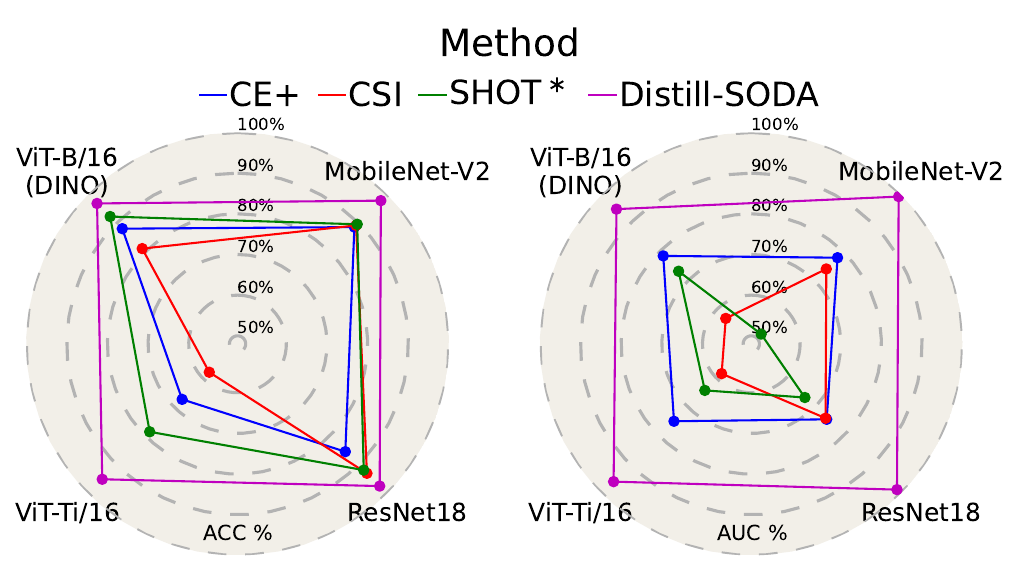}
    \caption{\textbf{Model architecture sensitivity} on Kather-16 $\to$ Kather-19
    adaptation (Split 1). Comparison of three source model architectures: MobileNet-V2, ResNet-18, ViT-Ti/16 and ViT-B/16 pre-trained with DINO in terms of {ACC$\uparrow$} and {AUC$\uparrow$}.}
    \label{fig:ablation_arch}
\end{figure}

\vspace{0.5em}
\noindent
\textbf{ViT self-training and style-based data augmentation.}
We evaluated the effect of our proposed style-based augmentation (AdvStyle) (Section \ref{subsec:ssl_adv_style}) for self-supervised training of ViT against the standard augmentation policies by comparing the quality of final pseudo-labels. Fig. \ref{fig:ablation_self_training}(a) shows that our AdvStyle policies significantly outperform conventional augmentation policies used in DINO \cite{caron2021emerging}, i.e., random color-jittering, and those utilized in T3PO \cite{galdran2022test} on every closed/open-set split on the Kather-16$\to$Kather-19 adaptation. Furthermore, Fig. \ref{fig:ablation_self_training}(b) illustrates the superior performance of our self-training strategy in comparison to solely relying on the pre-trained DINO weights without additional self-training for both ViT-B/16 and XCiT-S12/16 \cite{ali2021xcit} architectures.

\section{Practicality of $\ours$}
This section discusses the practicality of $\ours$ for clinical feasibility regarding source model architecture sensitivity, amount of target data, and runtime. As shown in Fig. \ref{fig:1}, SF-OSDA methods require two inputs: a pre-trained source model $f_s$ and unlabeled data from the target domain. We argue the source model architecture should be unknown a priori; therefore, SF-OSDA methods must be insensitive to model architecture. The target data's quantity is variable and may affect the SF-OSDA performance and computational cost. So, knowing the minimum amount of target data the model requires to be competitive with the baselines regarding runtime and performance is crucial. In the following, we evaluate $\ours$ under these two aspects. All experiments are conducted for Kather-16 $\to$ Kather-19 adaptation.
% on the three splits of Kather-16 $\to$ Kather-19.
%
%\Guillaume{In this section, we discuss the practicality of $\ours$ for clinical applications. SF-OSDA methods generally have two inputs: a source model $f_s$ and unlabeled data from the target domain. For the practical deployment of any SF-OSDA method, it should be insensitive to the architecture of the source model as well as the size of the target dataset. Here, we measure $\ours$'s sensitivity to different source model architectures, and also explicitly quantify the minimum amount of data needed without sacrificing the adaptation performance.
% In the following, we evaluate $\ours$ under these two axes. All experiments are conducted on Kather-16 $\to$ Kather-19 (\textbf{Split 1, 2 and 3}).}
%
%}

\vspace{0.5em}
\noindent
\textbf{Source model architecture sensitivity.}
We assess our method using different source model architectures, including MobileNet-V2, DINO ViT-B/16, ViT-Ti/16 \cite{touvron2021training}, and ResNet18 \cite{he2016deep} following the same training and adaptation procedure for each split. In Fig. \ref{fig:ablation_arch}, we compare $\ours$'s results against CE+, CSI, and \rev{SHOT$^\ast$} methods. We empirically show that our approach significantly surpasses all the baselines regarding ACC and AUC metrics for all three model architectures. Moreover, $\ours$'s performance seems invariant to the model architecture, an essential requirement for practicality.

\begin{figure}[t]
    \centering
    \includegraphics[width=\linewidth]{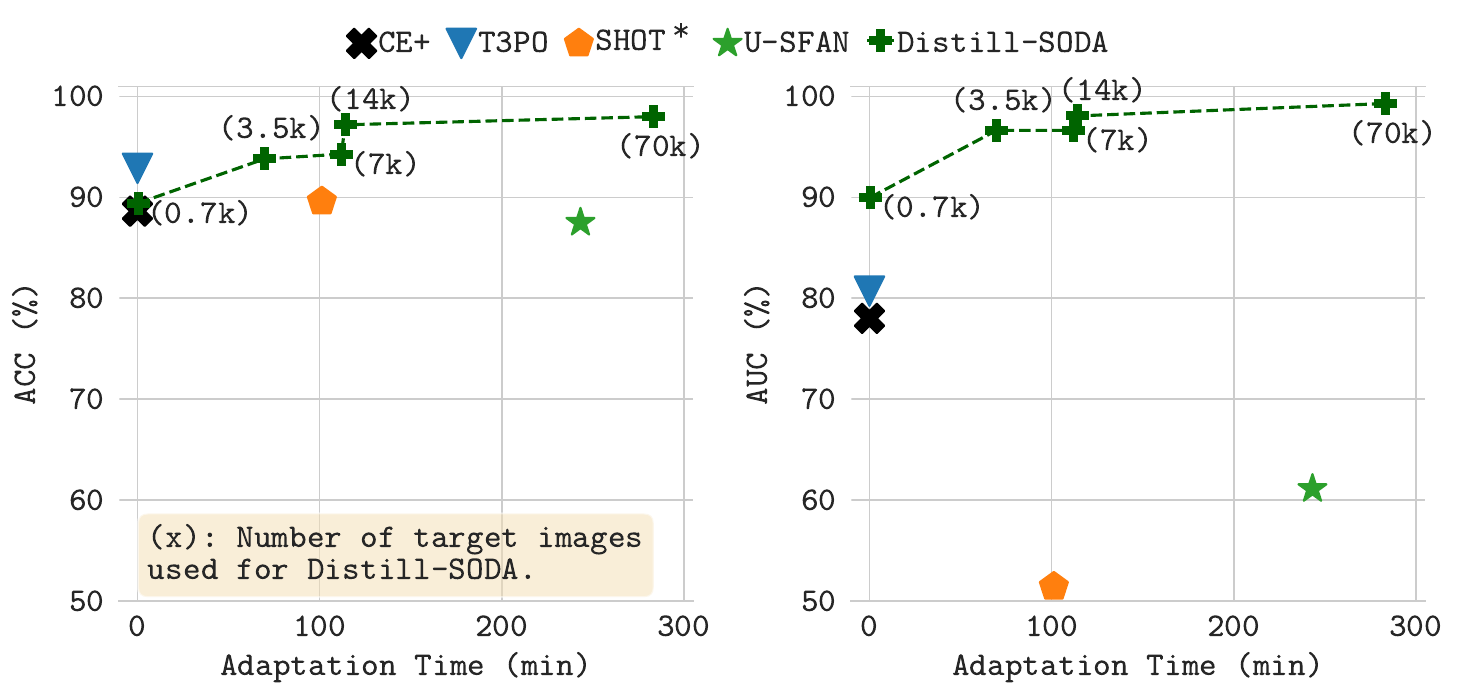}
    \caption{\textbf{Evaluating efficiency and data scalability:} The plot delineates the relationship between the performance of $\ours$ (measured by ACC/AUC metrics during the adaptation from Kather-16 $\to$ Kather-19, Split 1) and its operational time while adjusting for different volumes of target data.  Additionally, it compares our method with other methods that employ the entire target dataset comprising 70k images.}
    % We plot performance comparison in terms of ACC/AUC metrics on Kather-16 $\to$ Kather-19 adaptation (Split 1) by comparing the runtime and varying amount of target data needed by $\ours$ against open-set detection along with TTA and SF-OSDA methods that utilize full target dataset (70k images).}
    % \caption{\Devavrat{\textbf{Performance vs runtime and dataset size} on Kather-16 $\to$ Kather-19 (Split 1) adaptation. We plot runtime and amount of target data $\ours$ requires to attain certain ACC/AUC and compare it with open-set detection (CE+, T3PO); TTA (BN, OptTTA); and SF-OSDA (SHOT, U-SFAN) methods adapted on all 70k images.}}
    \label{fig:runtime}
\end{figure}

\vspace{0.5em}
\noindent
% \textbf{PerfromaAmount of target data for adaptation}
\textbf{Comparing performance, runtime, and dataset size.}
Fig. \ref{fig:runtime} shows the runtime of $\ours$ for adapting the source model on various target dataset sizes, along with their ACC and AUC scores. Our method achieves superior performance by using only 3.5k (5\%) unlabeled target domain images, outperforming other SF-OSDA baselines that utilize all 70k (100\%) target domain images. This shows the practical significance of $\ours$ with limited data availability, ensuring moderate processing times while cutting down on computational expenses for adaptation.

%Note that T3PO learns an auxiliary task of augmentation prediction during source training to detect open-set categories, while BN updates batch-normalization statistics of the source model with target domain images. However, these methods underperform under severe domain shifts.

% \Guillaume{We require nearly 4.7 hours of two NVIDIA GeForce GTX 1080 Ti GPU to adapt a source model $f_s$ on 70k image patches of Kather-19. In comparison, other SF-OSDA adaptation methods require 1.7 (SHOT) to 4 (U-SFAN) hours on the same amount of data. In Fig. \ref{fig:runtime}, we show we can reduce the adaptation time of our method to be competitive with the baselines by reducing the amount of target data without significant performance drops. For instance, $\ours$ only requires 1 hour adaptation time on 3.5k Kather-19 images (5\% of the total) and still gets significantly higher performance than all the baselines. Even though we can discuss T3PO is competitive with our method in terms of accuracy while not requiring any adaptation, we show AUC significantly benefits from our strategy even at the lowest computational cost. Additionally, the adaptation is required only once on the target data. It must only be repeated every time we face a new domain shift. Consequently, being practical in a low data regime, $\ours$ requires low computational cost and can be deployable for clinical applications.}

% source training strategies, source model architectures, target domain size, adversarial style augmentations for self-training on the target domain, hyper-parameter sensitivity, and self-distillation loss function.}
\section{Conclusion}
% We presented a novel source-free open-set domain adaptation method, $\ours$,  which enhances model robustness to simultaneous semantic and covariate shifts and disregards clinically irrelevant image regions from histological slides. 
% Our key contribution is distilling knowledge from a self-supervised vision transformer trained using our proposed adversarial style augmentations in the target domain.
We introduced $\ours$, a novel approach in source-free open-set domain adaptation designed to boost model resilience against both semantic and covariate shifts for histopathological image analysis. The core innovation lies in the knowledge transfer from a self-supervised vision transformer to the source model, specifically tailored to adapt when faced with open-set scenarios in the target domain. We have demonstrated that $\ours$ is not only compatible with existing pre-trained vision transformers but also excels when employed with public histopathology or specific target datasets. Additionally, we propose a novel style-based adversarial augmentation strategy that further refines the self-training of vision transformers directly on the target dataset.

Our approach has consistently outperformed multiple state-of-the-art baselines, especially in the context of three colorectal tissue datasets, confirming its robust efficacy. The design of $\ours$ facilitates seamless incorporation into an array of existing open-set detection methodologies, thereby bolstering their overall performance. Through meticulous ablation studies, we have illustrated the critical role played by each component of $\ours$, reinforcing the system's relevance and adaptability in clinical settings. Crucially, our method demonstrates admirable versatility by delivering reliable performance across diverse source model architectures. This versatility ensures that $\ours$ is not only theoretically sound but also practically valuable in clinical applications.

The primary limitation of $\ours$ is its reliance on having offline access to target images, which consequently limits its utility in scenarios that demand real-time, online adaptive capabilities. This gap underscores a vital area for future exploration: the expansion of $\ours$ to accommodate online settings. Our future work is firmly directed toward unraveling this challenge, aspiring to create an iteration of $\ours$ that brings forth the promise of real-time adaptability in the ever-changing landscape of medical image analysis.

%\begin{itemize}
%``data''

%\item \emph{Basic format for patents}$:$\\
%J. K. Author, ``Title of patent,'' U.S. Patent \emph{x xxx xxx}, Abbrev. Month, day, year.\\
%See \cite{b24}.

%%%%%%%%% REFERENCES
{\small
\bibliography{tmi}
\bibliographystyle{IEEEtran}
% \bibliography{egbib}
}
\end{document}